\documentclass[journal]{IEEEtran}
\usepackage{booktabs}
\usepackage{amsmath}
\usepackage{amssymb}
\usepackage{bm}
\usepackage{algorithm}
\usepackage{algorithmic}
\usepackage{color}
\usepackage{multirow}
\usepackage{subfig}
\usepackage{graphicx}
\usepackage{url}
\usepackage{mathtools}
\graphicspath{{./figs/}} 
\newcommand{\argmax}{\mathop{\rm argmax}\limits}

\newcommand{\Tref}[1]{Table~\ref{#1}}
\newcommand{\Eref}[1]{Eq.~(\ref{#1})}
\newcommand{\Fref}[1]{Fig.~\ref{#1}}
\newcommand{\Sref}[1]{Section~\ref{#1}}
\newcommand{\etal}[0]{et~al.}

\begin{document}
\title{PixelRL: Fully Convolutional Network with Reinforcement Learning for Image Processing}

\author{Ryosuke~Furuta,~\IEEEmembership{Member,~IEEE,}
        Naoto~Inoue,~\IEEEmembership{Student Member,~IEEE,}
        and~Toshihiko~Yamasaki,~\IEEEmembership{Member,~IEEE}
\thanks{R. Furuta is with the Department
of Information and Computer Technology, Tokyo University of Science,
Tokyo, 125-8585 Japan (e-mail: rfuruta@rs.tus.ac.jp). This work was done when R. Furuta was a Ph.D. student in The University of Tokyo.}
\thanks{N. Inoue and T. Yamasaki are with the Department of Information and Communication Engineering, Graduate School of Information Science and Technology, The University of Tokyo, Tokyo, 113-8656 Japan (e-mail: inoue@hal.t.u-tokyo.ac.jp; yamasaki@hal.t.u-tokyo.ac.jp).}
\thanks{Manuscript received April **, 2019; revised August **, 2019.}}

\markboth{IEEE Transactions on Multimedia,~Vol.~*, No.~*, ***~2019}%
{Furuta \MakeLowercase{\textit{et al.}}: PixelRL: Fully Convolutional Network with Multi-Step Reinforcement Learning for Image Processing}

\maketitle

\begin{abstract}
This paper tackles a new problem setting: reinforcement learning with pixel-wise rewards ({\it pixelRL}) for image processing.
After the introduction of the deep Q-network, deep RL has been achieving great success.
However, the applications of deep reinforcement learning (RL) for image processing are still limited.
Therefore, we extend deep RL to pixelRL for various image processing applications.
In pixelRL, each pixel has an agent, and the agent changes the pixel value by taking an action.
We also propose an effective learning method for pixelRL that significantly improves the performance by considering not only the future states of the own pixel but also those of the neighbor pixels.
The proposed method can be applied to some image processing tasks that require pixel-wise manipulations, where deep RL has never been applied.
Besides, it is possible to visualize what kind of operation is employed for each pixel at each iteration, which would help us understand why and how such an operation is chosen.
We also believe that our technology can enhance the explainability and interpretability of the deep neural networks.
In addition, because the operations executed at each pixels are visualized, we can change or modify the operations if necessary.

We apply the proposed method to a variety of image processing tasks: image denoising, image restoration, local color enhancement, and saliency-driven image editing.
Our experimental results demonstrate that the proposed method achieves comparable or better performance, compared with the state-of-the-art methods based on supervised learning.
The source code is available on \url{https://github.com/rfuruta/pixelRL}.
\end{abstract}


%
\IEEEpeerreviewmaketitle

\section{Introduction}

\IEEEPARstart{I}{n}
recent years, deep learning has been achieving great success not only in image classification, but also in image processing tasks such as image filtering, colorization, generation, and translation.
Although all of the above employ neural networks to learn the relationships between the input and output, their structures are approximately divided into two categories.
The neural networks typically used for image recognition are convolutional neural networks (CNNs) that have a number of convolution layers followed by fully-connected layers to output the classification scores.
In contrast, fully convolutional networks (FCNs) that do not have fully connected layers are employed for image processing tasks because a pixel-wise output is required for their tasks.

After the introduction of the deep Q-network (DQN)~\cite{mnih2013playing}, which can play Atari games on the human level, a lot more attention has been focused on deep reinforcement learning (RL).
Recently, deep RL has also been applied to a variety of image processing tasks~\cite{li2017a2,shuyue2018ffnet,jongchan2018distort}.
However, these methods can execute only global actions for the entire image and are limited to simple applications, e.g., image cropping~\cite{li2017a2} and global color enhancement~\cite{jongchan2018distort,hu2017exposure}.
These methods are hard to apply to applications that require pixel-wise manipulations such as image denoising.

To overcome this drawback, we propose a new problem setting: pixelRL for image processing.
PixelRL is a multi-agent RL problem, where each pixel has an agent.
The agents learn the optimal behavior to maximize the mean of the expected total rewards at all pixels.
Each pixel value is regarded as the current state and is iteratively updated by the agent's action.
Naively applying the existing techniques of the multi-agent RL to pixelRL is impractical in terms of computational cost because the number of agents is extremely large (e.g., 1 million agents for 1000x1000 pix images).
Therefore, we solve the problem by employing the fully convolutional network (FCN).
The merit of using FCN is that all the agents can share the parameters and learn efficiently.
In this paper, we also propose {\it reward map convolution}, which is an effective learning method for pixelRL.
By the proposed reward map convolution, each agent considers not only the future states of its own pixel but also those of the neighbor pixels.
Although the actions must be pre-defined for each application, the proposed method is interpretable by observing the actions executed by the agents, which is a novel and different point from the existing deep learning-based image processing methods for such applications.

The proposed pixelRL is applied to image denoising, image restoration, local color enhancement, and saliency-driven image editing.
To the best of our knowledge, this is the first work to apply RL to such low-level image processing for each pixel or each local region.
Our experimental results show that the agents trained with the pixelRL and the proposed reward map convolution achieve comparable or better performance, compared with state-of-the-art fully-supervised CNN-based methods.

The fundamental algorithm of the proposed method and the experimental results of the image denoising, image restoration, and local color enhancement have already been presented in our preliminary study~\cite{furuta2019fully}.
This paper provides a more detailed explanation and a new application of the pixelRL to saliency-driven image editing.

Our contributions are summarized as follows:
\begin{itemize}
\item We propose a novel problem setting: pixelRL for image processing, where the existing techniques for multi-agents RL cannot be applied. 
\item We propose {\it reward map convolution}, which is an effective learning method for pixelRL and boosts the performance.
\item We apply the pixelRL to image denoising, image restoration, local color enhancement, and saliency-driven image editing.
The proposed method is a completely novel approach for these tasks, and shows better or comparable performance, compared with state-of-the-art methods.
\item The actions executed by the agents are interpretable to humans, which is of great difference from conventional CNNs.
\end{itemize}

The rest of this paper is organized as follows. 
\Sref{sec:rel} reviews related works for deep RL and the four applications of pixelRL and discusses the differences between the proposed method and them.
\Sref{sec:back} briefly reviews the background knowledge of A3C~\cite{mnih2016asynchronous}, which is the base of the proposed method.
\Sref{sec:pixelRL} formulates the proposed pixelRL and describes our solution for the problem.
\Sref{sec:rconv} presents the reward map convolution to boosts the performance of the proposed method.
\Sref{sec:exp} presents the details of applications and experimental results.
Finally, \Sref{sec:conclusions} concludes this paper.

\section{Related Works}\label{sec:rel}
\subsection{Deep Reinforcement Learning}
Inspired by the success of the DQN~\cite{mnih2013playing}, a number of deep RL algorithms have been proposed~\cite{mnih2016asynchronous,schulman2015trust,nachum2017bridging}.
The primary application of deep RL is in playing video games such as Atari games, where the scores in the games are compared as the performance of the RL algorithm.

Here, we review the works that applied deep RL to computer vision and graphic tasks.
Caicedo and Lazebnik~\cite{caicedo2015active} and Mathe~\etal~\cite{mathe2016reinforcement} applied deep RL for object localization.
Their agents iteratively refine the bounding boxes that surrounded the objects in the input image as sequential actions.
Jie~\etal~\cite{jie2016tree} proposed a similar method based on the tree-structured approach to widely search objects in the input image.
Kong~\etal~\cite{kong2017collaborative} extended the method proposed by Caicedo and Lazebnik~\cite{caicedo2015active} to collaborative RL, where multiple agents detect multiple objects under an interaction by passing messages to each other.
The object localization technique based on deep RL is also used for object tracking in videos~\cite{xiang2015learning,yun2017action,supancic2017tracking}. 
Given the current frame and localized object as the current state, their agents localize the objects in the next frame as their actions, where that process is regarded as the Markov decision process.
Ganin~\etal~\cite{ganin2018synthesizing} proposed the SPIRAL agent that reconstructs the input image as a sequence of strokes (actions) on a graphics engine.
Rao~\etal~\cite{rao2017attention} applied deep RL to the frame selection for face recognition in videos.
The agent decides whether to drop each frame to maximize the feature distance between the videos of different persons, which is regarded as a hard attention model.
Lan~\etal~\cite{shuyue2018ffnet} proposed a video fast-forwarding method using deep RL, where the agent decides the number of frames to be skipped from the current frame.
Deep RL was also applied to weakly-supervised video summarization~\cite{chen2019weakly} in which hierarchical RL was proposed: the first-layer for choosing candidate shots and the second-layer for giving importance scores to each shot.

Very recently, deep RL has been used for some image processing applications.
Cao~\etal~\cite{cao2017attention} proposed a super-resolution method for face images.
The agent first chooses a local region and inputs it to the local enhancement network.
The enhancement network converts the local patch to a high-resolution one, and the agents choose the next local patch that should be enhanced.
This process is repeated until the maximum time step; consequently, the entire image is enhanced.
Li~\etal~\cite{li2017a2} used deep RL for image cropping.
The agent iteratively reshapes the cropping window to maximize the aesthetics score of the cropped image.
Yu~\etal~\cite{yu2018crafting} proposed the RL-restore method, where the agent selects a toolchain from a toolbox (a set of light-weight CNNs) to restore a corrupted image.
Park~\etal~\cite{jongchan2018distort} proposed a color enhancement method using DQN.
The agent iteratively chooses the image manipulation action (e.g., increase brightness) and retouches the input image.
The reward is defined as the negative distance between the retouched image by the agent and the one by an expert.
A similar idea is proposed by Hu~\etal~\cite{hu2017exposure}, where the agent retouches from RAW images.
As discussed in the introduction, all the above methods execute global actions for entire images.
In contrast, we tackle pixelRL, where pixel-wise actions can be executed.

Wulfmeier~\etal~\cite{wulfmeier2015maximum} used the FCN to solve the inverse reinforcement learning problem.
This problem setting is different from ours because one pixel corresponds to one state, and the number of agents is one in their setting.
In contrast, our pixelRL has one agent at each pixel.

\subsection{Image Denoising}
Image denoising methods can be classified into two categories: non-learning and learning based methods.
Many classical methods are categorized into the former class  (e.g., BM3D~\cite{dabov2007image}, non-local means filter~\cite{buades2005non}, and total variation minimization~\cite{rudin1992nonlinear}).
Although learning-based methods include dictionary-based methods such as~\cite{mairal2009non}, the recent trends in image denoising is neural network-based methods~\cite{xie2012image,burger2012image,zhang2017learning,lefkimmiatis2017non}.
Generally, neural-network-based methods have shown better performances, compared with non-leaning-based methods.

Our denoising method based on pixelRL is a completely different approach from the other neural network-based methods.
While most of neural-network-based methods learn to regress noise or true pixel values from a noisy input, our method iteratively removes noise with the sequence of simple pixel-wise actions (basic filters).

\subsection{Image Restoration}
Similar to image denoising, image restoration (also called image inpainting) methods are divided into non-learning and learning-based methods.
In the former methods such as~\cite{bertalmio2000image,criminisi2004region,bertalmio2003simultaneous}, the target blank regions are filled by propagating the pixel values or gradient information around the regions.
The filling process is highly sophisticated, but they are based on handcrafted algorithms.
Roth and Black~\cite{roth2005fields} proposed a Markov random field-based model to learn the image prior to the neighbor pixels.
Mairal~\etal~\cite{mairal2008sparse} proposed a learning-based method that creates a dictionary from an image database using K-SVD, and applied it to image  denoising and inpainting.
Recently, deep-neural-network-based methods were proposed~\cite{xie2012image,ren2015shepard,liu2019deep}, and the U-Net-based inpainting method~\cite{liu2019deep} showed much better performance than other methods.

Our method is categorized into the learning-based method because we used training images to optimize the policies.
However, similar to the classical inpainting methods, our method successfully propagates the neighbor pixel values with the sequence of basic filters.

\subsection{Color Enhancement}
One of the classical methods is color transfer proposed by Reinhard~\etal~\cite{reinhard2001color}, where the global color distribution of the reference image is transferred to the target image.
Hwang~\etal~\cite{hwang2012context} proposed an automatic local color enhancement method based on image retrieval.
This method enhances the color of each pixel based on the retrieved images with smoothness regularization, which is formulated as a Gaussian MRF optimization problem.

Yan~\etal~\cite{yan2016automatic} proposed the first color enhancement method based on deep learning.
They used a DNN to learn a mapping function from the carefully designed pixel-wise features to the desired pixel values.
Gharbi~\etal~\cite{gharbi2017deep} used a CNN as a trainable bilateral filter for high-resolution images and applied it to some image processing tasks.
Similarly, for fast image processing, Chen~\etal~\cite{chen2017fast} adopted an FCN to learn an approximate mapping from the input to the desired images.
Unlike deep learning-based methods that learn the input for an output mapping, our color enhancement method is interpretable because our method enhances each pixel value iteratively with actions such as~\cite{jongchan2018distort,hu2017exposure}.

\subsection{Saliency-Driven Image Editing}
Saliency-driven image editing is a task also called saliency retargeting~\cite{wong2011saliency}, image re-attentioning~\cite{nguyen2013image}, attention retargeting~\cite{mateescu2014attention}, and so on.
Given an input image and target regions (e.g., a mask image), the input image is edited such that the saliency values in the target regions increase/decrease.
Its objective is to enhance/diminish the target objects in the input image naturally.
Most of the existing methods are based on hand-crafted algorithms (e.g., saliency values are controlled by changing texture~\cite{Su:05:Texture}, color~\cite{mateescu2014attention,hagiwara2011saliency}, orientation~\cite{mateescu2013guiding}, luminance, saturation, and sharpness~\cite{wong2011saliency}).
Nguyen~\etal~\cite{nguyen2013image} proposed a method based on MRF optimization to keep the smoothness between the enhanced regions and the neighbor regions.
Recently, Mechrez~\etal~\cite{mechrez2018saliency} made a benchmark for this task and showed that their proposed method achieved the best performance.
This task is also applied to 3D volume~\cite{kim2006saliency} and 3D mesh~\cite{kim2008persuading}.

Our method is different from the existing methods because the agents learn how to edit the input image in order to increase/decrease the saliency values.
This is the first learning-based method for saliency-driven image editing task.

\section{Background Knowledge}\label{sec:back}
In this paper, we extend the asynchronous advantage actor-critic (A3C)~\cite{mnih2016asynchronous} for the pixelRL problem because A3C showed good performance with efficient training in the original paper\footnote{Note that we can employ any deep RL methods such as DQN instead of A3C.}.
In this section, we briefly review the training algorithm of A3C.
A3C is one of the actor-critic methods, which has two networks: policy network and value network.
We denote the parameters of each network as $\theta_p$ and $\theta_v$, respectively.
Both networks use the current state $s^{(t)}$ as the input, where $s^{(t)}$ is the state at the time step $t$.
The value network outputs the value $V(s^{(t)})$: the expected total rewards from state $s^{(t)}$, which shows how good the current state is.
The gradient for $\theta_v$ is computed as follows:
\begin{multline}
R^{(t)}=r^{(t)}+\gamma r^{(t+1)}+\gamma^2 r^{(t+2)}+\cdots \\
+\gamma^{n-1}r^{(t+n-1)}+\gamma^n V(s^{(t+n)}),\label{eq:R}
\end{multline}
\begin{equation}
d\theta_v=\nabla_{\theta_v}\left(R^{(t)}-V(s^{(t)})\right)^2,
\end{equation}
where $\gamma^i$ is the $i$-th power of the discount factor $\gamma$.

The policy network outputs the policy $\pi(a^{(t)}|s^{(t)})$ (probability through softmax) of taking action $a^{(t)}\in \mathcal{A}$.
Therefore, the output dimension of the policy network is $|\mathcal{A}|$.
The gradient for $\theta_p$ is computed as follows:
\begin{eqnarray}
A(a^{(t)},s^{(t)})&=&R^{(t)}-V(s^{(t)}),\label{eq:original_adv} \\ 
d\theta_p&=&-\nabla_{\theta_p}\log \pi(a^{(t)}|s^{(t)})A(a^{(t)},s^{(t)}).
\end{eqnarray}
$A(a^{(t)},s^{(t)})$ is called the advantage, and $V(s^{(t)})$ is subtracted in~\Eref{eq:original_adv} to reduce the variance of the gradient.
For more details, see~\cite{mnih2016asynchronous}.

\section{Reinforcement Learning with Pixel-wise Rewards (PixelRL)}\label{sec:pixelRL}
Here, we describe the proposed pixelRL problem setting.
Let $I_i$ be the $i$-th pixel in the input image $\bm{I}$ that has $N$ pixels $(i=1,\cdots,N)$.
Each pixel has an agent, and its policy is denoted as $\pi_i(a_i^{(t)}|s_i^{(t)})$, where $a_i^{(t)} (\in \mathcal{A})$ and $s_i^{(t)}$ are the action and the state of the $i$-th agent at the time step $t$, respectively.
$\mathcal{A}$ is the pre-defined action set, and $s_i^{(0)}=I_i$. 
The agents obtain the next states $\bm{s}^{(t+1)}=(s_1^{(t+1)},\cdots,s_N^{(t+1)})$ and rewards $\bm{r}^{(t)}=(r_1^{(t)},\cdots,r_N^{(t)})$ from the environment by taking the actions $\bm{a}^{(t)}=(a_1^{(t)},\cdots,a_N^{(t)})$.
The objective of the pixelRL problem is to learn the optimal policies $\bm{\pi}=(\pi_1,\cdots,\pi_N)$ that maximize the mean of the total expected rewards at all pixels:
\begin{eqnarray}
\bm{\pi}^*&=&\argmax_{\bm{\pi}}E_{\bm{\pi}}\left(\sum_{t=0}^{\infty}\gamma^t \overline{r}^{(t)}\right),\\
\overline{r}^{(t)}&=&\frac{1}{N}\sum_{i=1}^N r^{(t)}_i,
\end{eqnarray}
where $\overline{r}^{(t)}$ is the mean of the rewards $r_i^{(t)}$ at all pixels.

A naive solution for this problem is to train a network that outputs Q-values or policies for all possible set of actions $\bm{a}^{(t)}$.
However, it is computationally impractical because the dimension of the last fully connected layer must be $|\mathcal{A}|^N$, which is too large.

\begin{figure}[t]
    \begin{center}
        \includegraphics[width=1.0\linewidth]{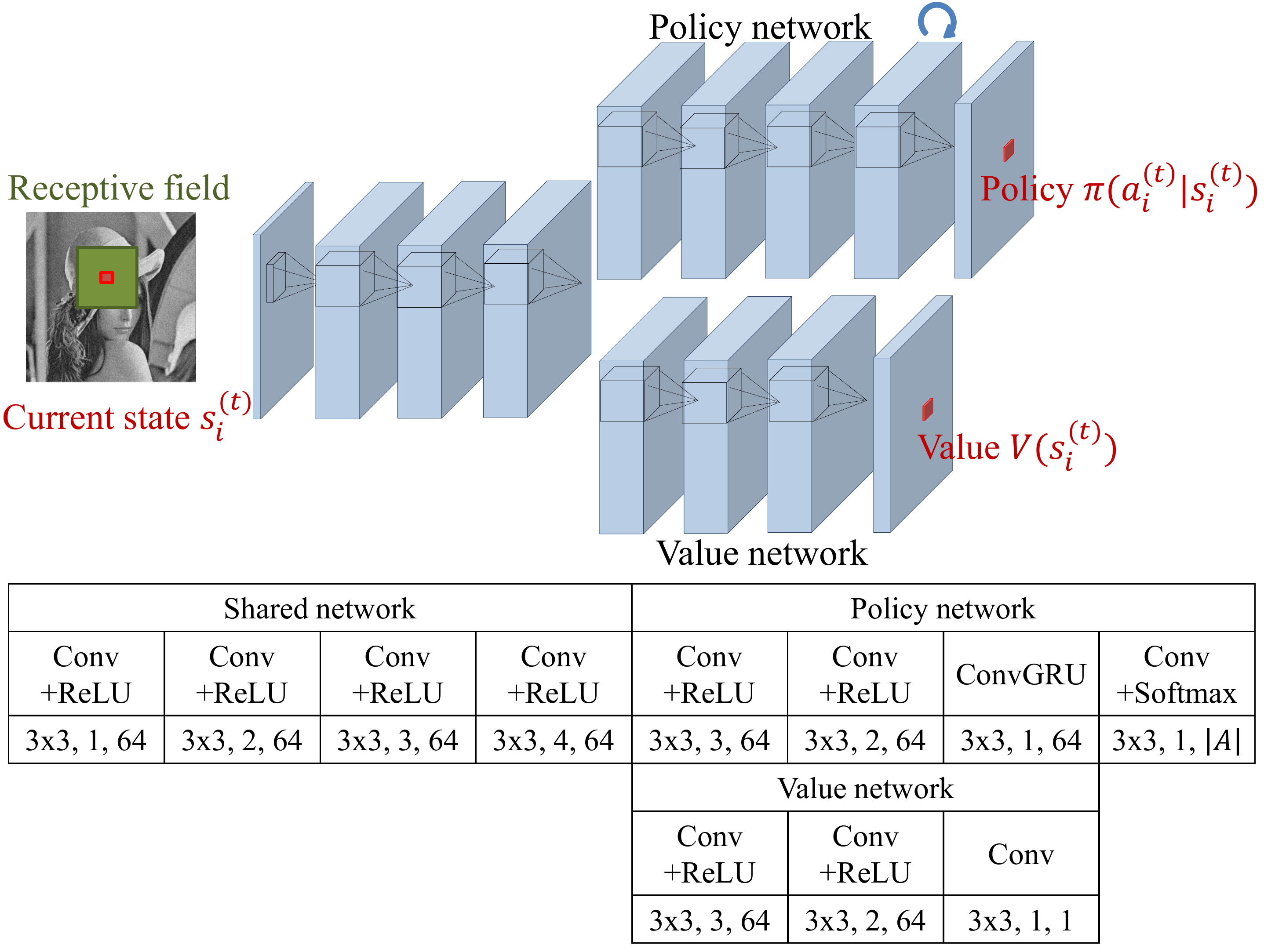}
        \caption{Network architecture of the fully convolutional A3C. The numbers in the table denote the filter size, dilation factor, and output channels, respectively.}
        \label{fig:arch}
    \end{center}    
\end{figure}

Another solution is to divide this problem into $N$ independent subproblems and train $N$ networks, where we train the $i$-th agent to maximize the expected total reward at the $i$-th pixel:
\begin{equation}
\pi_i^*=\argmax_{\pi_i}E_{\pi_i}\left(\sum_{t=0}^{\infty}\gamma^t r_i^{(t)} \right).
\end{equation}
However, training $N$ networks is also computationally impractical when the number of pixels is large.
In addition, it treats only the fixed size of images.
To solve the problems, we employ a FCN instead of $N$ networks.
By using the FCN, all the $N$ agents can share the parameters, and we can parallelize the computation of $N$ agents on a GPU, which renders the training efficient.
In this paper, we employ A3C and extend it to the fully convolutional form.
We illustrate its architecture in~\Fref{fig:arch}.

The pixelRL setting is different from typical multi-agent RL problems in terms of two points.
The first point is that the number of agents $N$ is extremely large ($>10^5$).
Therefore, typical multi-agent learning techniques such as~\cite{lowe2017multi} cannot be directly applied to the pixelRL.
Next, the agents are arrayed in a 2D image plane.
In the next section, we propose an effective learning method that boosts the performance of the pixelRL agents by leveraging this property, named {\it reward map convolution}.

\section{Reward Map Convolution}\label{sec:rconv}
Here, for the ease of understanding, we first consider the one-step learning case (i.e., $n=1$ in~\Eref{eq:R}).

When the receptive fields of the FCNs are 1x1 (i.e., all the convolution filters in the policy and value network are 1x1), the $N$ subproblems are completely independent.
In that case, similar to the original A3C, the gradient of the two networks are computed as follows:
\begin{eqnarray}
R_i^{(t)}&=&r_i^{(t)}+\gamma V(s_i^{(t+1)}),\label{eq:ith_R}\\
d\theta_v&=&\nabla_{\theta_v}\frac{1}{N}\sum_{i=1}^N\left(R_i^{(t)}-V(s_i^{(t)})\right)^2, \label{eq:ith_gradv}\\
A(a_i^{(t)},s_i^{(t)})&=&R_i^{(t)}-V(s_i^{(t)}), \label{eq:i_adv}
\end{eqnarray}
\begin{equation}
d\theta_p=-\nabla_{\theta_p}\frac{1}{N}\sum_{i=1}^N\log \pi(a_i^{(t)}|s_i^{(t)})A(a_i^{(t)},s_i^{(t)}).\label{eq:ith_gradp}
\end{equation}
As shown in~Eqs.~(\ref{eq:ith_gradv}) and (\ref{eq:ith_gradp}), the gradient for each network parameter is the average of the gradients at all pixels.

However, one of the recent trends in CNNs is to enlarge the receptive field to boost the network performance~\cite{yu2017dilated,zhang2017learning}.
Our network architecture, which was inspired by~\cite{zhang2017learning} in~\Fref{fig:arch}, has a large receptive field.
In this case, the policy and value networks observe not only the $i$-th pixel $s_i^{(t)}$ but also the neighbor pixels to output the policy $\pi$ and value $V$ at the $i$-th pixel.
In other words, the action $a_i^{(t)}$ affects not only the state $s_i^{(t+1)}$ but also the policies and values in $\mathcal{N}(i)$ at the next time step, where $\mathcal{N}(i)$ is the local window centered at the $i$-th pixel.
Therefore, to consider it, we replace $R_i$ in~\Eref{eq:ith_R} as follows:
\begin{equation}
R_i^{(t)}=r_i^{(t)}+\gamma \sum_{j\in \mathcal{N}(i)}w_{i-j} V(s_j^{(t+1)}),\label{eq:ith_R_conv}
\end{equation}
where $w_{i-j}$ is the weight that means how much we consider the values $V$ of the neighbor pixels at the next time step ($t+1$).
$\bm{w}$ can be regarded as a convolution filter weight and can be learned simultaneously with the network parameters $\theta_p$ and $\theta_v$.
It is noteworthy that the second term in~\Eref{eq:ith_R_conv} is a 2D convolution because each pixel $i$ has a 2D coordinate $(i_x,i_y)$.

Using the matrix form, we can define the $\bm{R}^{(t)}$ in the $n$-step case.
\begin{multline}
\bm{R}^{(t)}=\bm{r}^{(t)}+\gamma \bm{w}*\bm{r}^{(t+1)}+\gamma^2 \bm{w}^2*\bm{r}^{(t+2)}+\cdots \\
+\gamma^{n-1}\bm{w}^{n-1}*\bm{r}^{(t+n-1)}+\gamma^n\bm{w}^n*\bm{V}(\bm{s}^{(t+n)}),\label{eq:nstep_R_conv}
\end{multline}
where $\bm{R}^{(t)}$, $\bm{r}^{(t)}$, and $\bm{V}(\bm{s}^{(t)})$ are the matrices whose $(i_x,i_y)$-th elements are ${R}_i^{(t)}$, ${r}_i^{(t)}$, and $V({s}_i^{(t)})$, respectively. $*$ is the convolution operator, and $\bm{w}^{n}*\bm{r}$ denotes the $n$-times convolution on $\bm{r}$ with the 2D convolution filter $\bm{w}$.
The gradients can be denoted as follows using the matrix form:
\begin{gather}
d\theta_v=\nabla_{\theta_v}\frac{1}{N}\bm{1}^\top\left\{\left(\bm{R}^{(t)}-\bm{V}(\bm{s}^{(t)})\right)\odot\left(\bm{R}^{(t)}-\bm{V}(\bm{s}^{(t)})\right)\right\}\bm{1}, \label{eq:mat_gradv}\\
\bm{A}(\bm{a}^{(t)},\bm{s}^{(t)})=\bm{R}^{(t)}-\bm{V}(\bm{s}^{(t)}), \label{eq:mat_adv}\\
d\theta_p=-\nabla_{\theta_p}\frac{1}{N}\bm{1}^\top\left\{\log \bm{\pi}(\bm{a}^{(t)}|\bm{s}^{(t)})\odot \bm{A}(\bm{a}^{(t)},\bm{s}^{(t)})\right\}\bm{1},\label{eq:mat_gradp}
\end{gather}
where $\bm{A}(\bm{a}^{(t)},\bm{s}^{(t)})$, and $\bm{\pi}(\bm{a}^{(t)}|\bm{s}^{(t)})$ are the matrices whose $(i_x,i_y)$-th elements are $A({a}_i^{(t)},{s}_i^{(t)})$, and $\pi({a}_i^{(t)}|{s}_i^{(t)})$, respectively.
$\bm{1}$ is an all-ones vector where every element is one, and $\odot$ denotes element-wise multiplication.

Similar to $\theta_p$ and $\theta_v$ in~Eqs.~(\ref{eq:ith_gradv}) and (\ref{eq:ith_gradp}), the gradient for $\bm{w}$ is computed as follows:
\begin{align}
\begin{multlined}[c]
d\bm{w}=-\nabla_{\bm{w}}\frac{1}{N}\sum_{i=1}^N\log\pi(a_i^{(t)}|s_i^{(t)})(R_i^{(t)}-V(s_i^{(t)}))\\
+\nabla_{\bm{w}}\frac{1}{N}\sum_{i=1}^N(R_i^{(t)}-V(s_i^{(t)}))^2,\label{eq:gradw}
\end{multlined}\\
\begin{multlined}[c]
\phantom{d\bm{w}}=-\nabla_{\bm{w}}\frac{1}{N}\bm{1}^\top\left\{\log \bm{\pi}(\bm{a}^{(t)}|\bm{s}^{(t)})\odot \bm{A}(\bm{a}^{(t)},\bm{s}^{(t)})\right\}\bm{1}\\
+\nabla_{\bm{w}}\frac{1}{N}\bm{1}^\top\left\{\left(\bm{R}^{(t)}-\bm{V}(\bm{s}^{(t)})\right)\odot\left(\bm{R}^{(t)}-\bm{V}(\bm{s}^{(t)})\right)\right\}\bm{1}.
\end{multlined}
\end{align}

Similar to typical policy gradient algorithms, the first term in~\Eref{eq:gradw} encourages a higher expected total reward.
The second term operates as a regularizer such that $R_i$ is not deviated from the prediction $V(s_i^{(t)})$ by the convolution.
We summarize the training algorithm of the fully convolutional A3C with the proposed reward map convolution in Algorithm~\ref{alg}.
The differences from the original A3C are highlighted in red.
\begin{algorithm*}[!t]
\caption{Training pseudo-code of fully convolutional A3C with the proposed reward map convolution}
\label{alg}
\begin{algorithmic}
\STATE // Assume global shared parameter vectors $\theta_p$, $\theta_v$, and \textcolor{red}{$\bm{w}$} and global counter $T=0$.
\STATE // Assume thread-specific parameter vectors $\theta'_p$, $\theta'_v$, and \textcolor{red}{$\bm{w}'$}.
\STATE Initialize thread step counter $t\leftarrow1$.
\REPEAT
\STATE Reset gradients: $d\theta_p \leftarrow 0$, $d\theta_v \leftarrow 0$, and \textcolor{red}{$d\bm{w} \leftarrow 0$}.
\STATE Synchronize thread-specific parameters $\theta_p'=\theta_p$, $\theta_v'=\theta_v$, and \textcolor{red}{$\bm{w}'=\bm{w}$}
\STATE $t_{start}=t$
\STATE Obtain state $s_i^{(t)}$ for $\forall i$
\REPEAT
\STATE Perform $a_i^{(t)}$ according to policy $\pi(a_i^{(t)}|s_i^{(t)})$ for $\forall i$
\STATE Receive reward $r_i^{(t)}$ and new state $s_i^{(t+1)}$ for $\forall i$
\STATE $t\leftarrow t+1$
\STATE $T\leftarrow T+1$
\UNTIL terminal $s_i^{(t)}$ or $t-t_{start}==t_{max}$
\STATE for $\forall i$ $R_i=\begin{cases}0 & \text{for terminal}\ s_i^{(t)} \\ V(s_i^{(t)}) & \text{for non-terminal}\ s_i^{(t)}\end{cases}$
\FOR{$k \in \{t-1,\cdots,t_{start}\}$}
\STATE $R_i \leftarrow \gamma R_i$
\STATE \textcolor{red}{Convolve $\bm{R}$ with $\bm{w}$: $R_i \leftarrow \sum_{j\in \mathcal{N}(i)}w_{i-j}R_j$ for $\forall i$}
\STATE $R_i \leftarrow r_i^{(k)}+R_i$
\STATE Accumulate gradients w.r.t. $\theta'_p$: $d\theta_p\leftarrow d\theta_p-\nabla_{\theta_p'}\textcolor{red}{\frac{1}{N}\sum_{i=1}^N}\log\pi(a_i^{(k)}|s_i^{(k)})(R_i-V(s_i^{(k)}))$
\STATE Accumulate gradients w.r.t. $\theta'_v$: $d\theta_v\leftarrow d\theta_v+\nabla_{\theta_v'}\textcolor{red}{\frac{1}{N}\sum_{i=1}^N}(R_i-V(s_i^{(k)}))^2$
\STATE \textcolor{red}{Accumulate gradients w.r.t. $\bm{w}'$: $d\bm{w}\leftarrow d\bm{w}-\nabla_{\bm{w}}\frac{1}{N}\sum_{i=1}^N\log\pi(a_i^{(k)}|s_i^{(k)})(R_i-V(s_i^{(k)}))+\nabla_{\bm{w}}\frac{1}{N}\sum_{i=1}^N(R_i-V(s_i^{(k)}))^2$}
\ENDFOR
\STATE Update $\theta_p$, $\theta_v$, and \textcolor{red}{$\bm{w}$} using $d\theta_p$, $d\theta_v$, and \textcolor{red}{$d\bm{w}$}, respectively.
\UNTIL $T>T_{max}$
\end{algorithmic}
\end{algorithm*}

\section{Applications and Results}\label{sec:exp}
We implemented the proposed method on Python with Chainer \cite{tokui2019chainer} and ChainerRL~\footnote{https://github.com/chainer/chainerrl} libraries, and applied it to four different applications.

\subsection{Image denoising}
\subsubsection{Method}
\begin{table}[t]
\caption{Actions for image denoising and restoration.}
\centering
{
{
  \begin{tabular}{cccc} \toprule
      & action & filter size & parameter \\ \toprule
     1 & box filter & 5x5 & - \\
     2 & bilateral filter & 5x5 & $\sigma_c=1.0, \sigma_S=5.0$ \\
     3 & bilateral filter & 5x5 & $\sigma_c=0.1, \sigma_S=5.0$ \\
     4 & median filter & 5x5 & - \\
     5 & Gaussian filter & 5x5 & $\sigma=1.5$ \\
     6 & Gaussian filter & 5x5 & $\sigma=0.5$ \\
     7 & pixel value += 1 & - & - \\
     8 & pixel value -= 1 & - & - \\
     9 & do nothing & - & - \\
 \bottomrule
  \end{tabular} 
}
}
\label{tbl:actions_denoise}
\end{table}
The input image $\bm{I}(=\bm{s}^{(0)})$ is a noisy gray scale image, and the agents iteratively remove the noises by executing actions. 
It is noteworthy that the proposed method can also be applied to color images by independently manipulating on the three channels.
In \Tref{tbl:actions_denoise}, we show the list of actions that the agents can execute, which were empirically decided.
Note that all the actions are classical image filtering algorithms.
We defined the reward $r_i^{(t)}$ as follows:
\begin{equation}
r_i^{(t)}=(I_i^{target}-s_i^{(t)})^2-(I_i^{target}-s_i^{(t+1)})^2,\label{eq:reward_denoise}
\end{equation}
where $I_i^{target}$ is the $i$-th pixel value of the original clean image.
Intuitively, \Eref{eq:reward_denoise} means how much the squared error on the $i$-th pixel was decreased by the action $a_i^{(t)}$.
As shown in~\cite{maes2009structured}, maximizing the total reward in~\Eref{eq:reward_denoise} is equivalent to minimizing the squared error between the final state $\bm{s}^{(t_{max})}$ and the original clean image $\bm{I}^{target}$.

\subsubsection{Implementation details}
We used BSD68 dataset~\cite{roth2005fields}, which has 428 train images and 68 test images.
Similar to~\cite{zhang2017learning}, we added 4,774 images from Waterloo exploration database~\cite{ma2017waterloo} to the training set.
We set the minibatch size to 64, and the training images were augmented with $70\times 70$ random cropping, left-right flipping, and random rotation.
To train the fully convolutional A3C, we used the ADAM optimizer~\cite{kingma2014adam} and the poly learning, where the learning rate started from $1.0\times 10^{-3}$ and multiplied by $(1-\frac{episode}{max\_episode})^{0.9})$ at each episode. 
We set the $max\_episode$ to 30,000 and the length of each episode $t_{max}$ to 5.
Therefore, the maximum global counter $T_{max}$ in Algorithm~\ref{alg} was $30,000 \times 5=150,000$.
To reduce the training time, we initialize the weights of the fully convolutional A3C with the publicly available weights of~\cite{zhang2017learning}, except for the convGRU and the last layers.
We adopted the stepwise training: the fully convolutional A3C was trained first, subsequently it was trained again with the reward map convolution.
We set the filter size of $\bm{w}$ to $33\times 33$, which is equal to the receptive field size of the networks in~\Fref{fig:arch}.
The number of asynchronous threads was one (i.e., equivalent to A2C: advantage actor-critic).
$\bm{w}$ was initialized as the identity mapping (i.e., only the center of $\bm{w}$ was one, and zero otherwise).
It required approximately 16 hours for the 30,000 episode training, and 0.44 sec on average for a test image whose size is $481\times 321$ on a single Tesla V100 GPU.

\subsubsection{Results}\label{sec:denoise_results}
\begin{table}[t]
\caption{PSNR [dB] on BSD68 test set with Gaussian noise.}
\centering
{
{
  \begin{tabular}{ccc|ccc} \toprule
     \multicolumn{3}{c|}{\multirow{2}{*}{Method}} & \multicolumn{3}{c}{std. $\sigma$}\\
      & & & 15 & 25 & 50 \\ \toprule
     \multicolumn{3}{c|}{BM3D~\cite{dabov2007image}} & 31.07 & 28.57 & 25.62 \\
     \multicolumn{3}{c|}{WNNM~\cite{gu2014weighted}} & 31.37 & 28.83 & 25.87 \\
     \multicolumn{3}{c|}{TNRD~\cite{chen2017trainable}} & 31.42 & 28.92 & 25.97 \\
     \multicolumn{3}{c|}{MLP~\cite{burger2012image}} & - & 28.96 & 26.03 \\
     \multicolumn{3}{c|}{CNN~\cite{zhang2017learning}} & 31.63 & 29.15 & 26.19 \\
     \multicolumn{3}{c|}{CNN~\cite{zhang2017learning} +aug.} & {\bf 31.66} & {\bf 29.18} & {\bf 26.20} \\ \midrule
     \multicolumn{3}{c|}{Original} & 24.79 & 20.48 & 14.91 \\
     \multicolumn{3}{c|}{Random Agents} & 24.69 & 24.30 & 22.80 \\ \midrule
     \multicolumn{3}{c|}{Proposed} & & & \\
     +convGRU & +RMC & +aug. & & & \\
      & & & 31.17 & 28.75 & 25.78 \\
      \checkmark& & & 31.26 & 28.83 & 25.87 \\
      \checkmark&\checkmark & & 31.40 & 28.85 & 25.88 \\
      \checkmark&\checkmark &\checkmark & 31.49 & 28.94 & 25.95 \\
 \bottomrule
  \end{tabular} 
}
}
\label{tbl:comp_gaussian}
\end{table}
In~\Tref{tbl:comp_gaussian}, we show the comparison of Gaussian denoising with other methods.
RMC is the abbreviation for reward map convolution.
``Aug.'' means the data augmentation for test images, where a single test image was augmented to eight images by a left-right flip and $90^{\circ}$, $180^{\circ}$, and $270^{\circ}$ rotations, similar to~\cite{timofte2017ntire}.
``Original'' means the PSNR between the input and groundtruth images.
As a baseline, we show the result of ``random agents,'' which randomly chose actions from~\Tref{tbl:actions_denoise} at each time step.
When the noise level is high ($\sigma=25$ and $50$), the random agents can obtain higher PSNR than original images because the six of nine actions in~\Tref{tbl:actions_denoise} are filtering actions designed for image denoising.
However, when the noise level is low ($\sigma=15$), the random agents cannot improve the PSNR from the original images.
In contrast, the proposed method successfully leaned a strategy for denoising and obtained much better results than the ``random agents'' baseline.
We observed that CNN~\cite{zhang2017learning} is the best.
However, the proposed method achieved the comparable results with the other state-of-the-art methods.
Adding the convGRU to the policy network improved the PSNR by approximately +0.1dB.
The RMC significantly improved the PSNR when $\sigma=15$, but improved little when $\sigma=25$ and $50$.
That is because the agents can obtain much reward by removing the noises at their own pixels rather than considering the neighbor pixels when the noises are strong.
The augmentation for test images further boosted the performance.
We report the CNN~\cite{zhang2017learning} with the same augmentation for a fair comparison.
It demonstrates that the pixel-wise combination of traditional image filtering methods can perform better than classical sophisticated denoising methods (BM3D~\cite{dabov2007image} and WNNM~\cite{gu2014weighted}) and comparative to state-of-the-art NN-based methods (TNRD~\cite{chen2017trainable}, MLP~\cite{burger2012image}, and CNN~\cite{zhang2017learning}).

\begin{figure}[t]
    \begin{center}
        \includegraphics[width=1.0\linewidth]{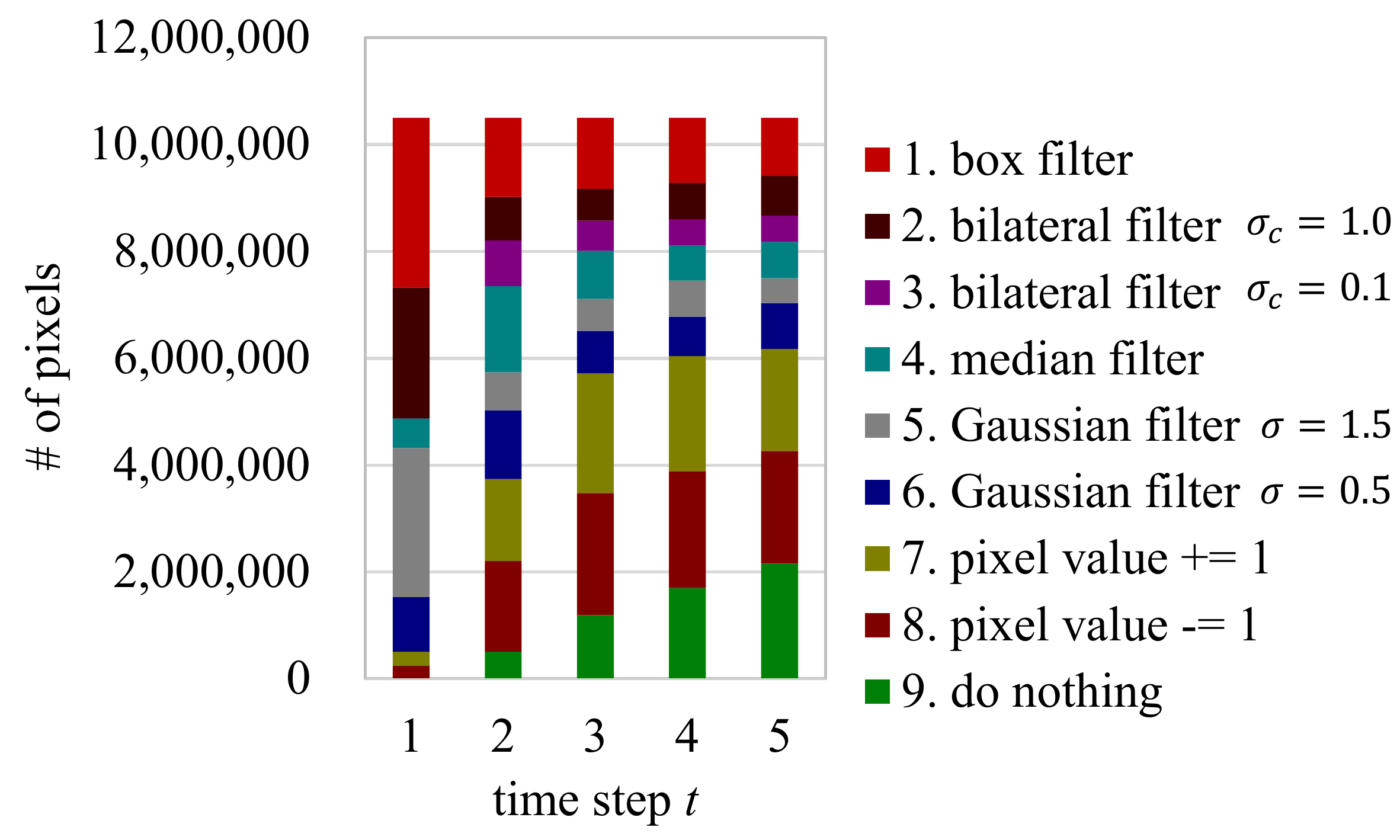}
        \caption{Number of actions executed at each time step for Gaussian denoising ($\sigma=50$) on the BSD68 test set.}
        \label{fig:actions_denoise}
    \end{center}    
\end{figure}
In \Fref{fig:actions_denoise}, we show the number of actions executed by the proposed method at each time step for Gaussian denoising ($\sigma=50$) on the BSD68 test set.
We observed that the agents successfully obtained a strategy in which they first removed the noises using strong filters (box filter, bilateral filter $\sigma_c=1.0$, and Gaussian filter $\sigma=1.5$); subsequently they adjusted the pixel values by the other actions (pixel values +=1 and -=1).

\begin{table}[t]
\caption{PSNR [dB] on Gaussian denoising with different action sets.}
\centering
{
  \begin{tabular}{c|ccc} \toprule
     \multirow{2}{*}{Actions} & \multicolumn{3}{c}{std. $\sigma$}\\
      & 15 & 25 & 50 \\ \toprule
     only basic filters & 29.82 & 27.60 & 25.20 \\
     + bilateral filter $\sigma_c=1.0$ & 29.93 & 27.81 & 25.30 \\
     + pixel value $\pm=$ 1 & 30.72 & 28.25 & 25.59 \\
     + different filter parameters. & 31.26 & 28.83 & 25.87 \\
 \bottomrule
  \end{tabular} 
}
\label{tbl:ablation_gaussian}
\end{table}
We conducted the ablation studies with different sets of actions on Gaussian denoising. 
We show the results in~\Tref{tbl:ablation_gaussian}.
When the actions were only basic filters ([1] box filters, [4] median filter, [5] Gaussian filter with $\sigma=1.5$, and [9] do nothing in Table 1 in our main paper), its PSNRs were 29.82, 27.60, and 25.20 for noise std. 15, 25, and 50, respectively. 
When we added [2] bilateral filter with $\sigma_c=1.0$, the PSNRs increased to 29.93, 27.81, and 25.30. 
In addition, when we added the two actions ([7] pixel value += 1 and [8] pixel value -= 1), the PSNRs further increased to 30.72, 28.25, and 25.59. 
Finally, when we added the two filters with different parameters ([3] bilateral filter with $\sigma_c=0.1$ and [6] Gaussian filter with $\sigma=0.5$), the PSNRs were 31.26, 28.83, and 25.87. 
Therefore, all the actions are important for the high performance although it may be further increased if we can find more proper action set.
Although we tried adding some advanced filters for image denoising such as guided filter~\cite{he2010guided} and non-local means filter~\cite{buades2005non}, the performance was not improved any more.

\begin{figure}[t]
    \begin{center}
        \includegraphics[width=1.0\linewidth]{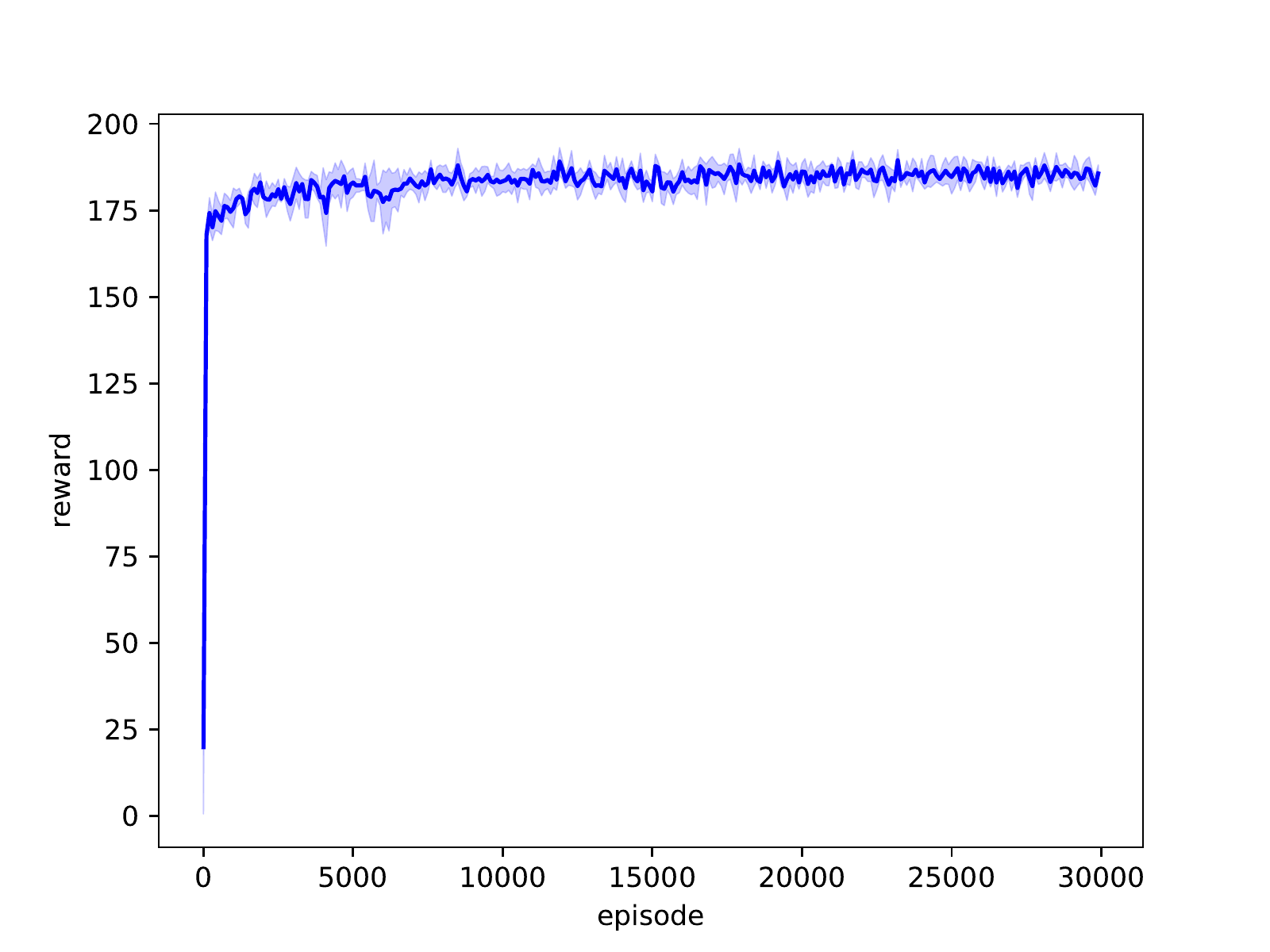}
        \caption{Mean and standard deviation of averaged accumulated rewards at each episode during training ($\sigma=15$) on five trials.}
        \label{fig:training_curve}
    \end{center}    
\end{figure}
We show the training curve (averaged accumulated rewards at each episode: $\hat{r}$ in~\Eref{eq:acc_reward}) of the proposed method in~\Fref{fig:training_curve}.
\begin{equation}
    \hat{r}=\frac{1}{N}\sum_{i=1}^N\sum_{t=0}^{t_{max}}\gamma^t r^{(t)}_i.\label{eq:acc_reward}
\end{equation}
The blue line and light blue area show the mean and standard deviation of $\hat{r}$ on five trials, respectively.
The training of the proposed method is stable, and the agents obtain a good strategy immediately because all the agents share the parameters and the gradients are averaged as shown in Eqs.~(\ref{eq:mat_gradv}) and (\ref{eq:mat_gradp}).
In other words, the agents can obtain $N$ ($>10^5$) experience at one backward pass in contrast to usual reinforcement learning where only one experience is obtained at one backward pass.

\begin{table}[t]
\caption{PSNR [dB] on BSD68 test set with Poisson noise.}
\centering
{
  \begin{tabular}{ccc|ccc} \toprule
     \multicolumn{3}{c|}{\multirow{2}{*}{Method}} & \multicolumn{3}{c}{Peak Intensity}\\
      & & & 120 & 30 & 10 \\ \toprule
     \multicolumn{3}{c|}{CNN~\cite{zhang2017learning}} & 31.62 & 28.20 & 25.93 \\
     \multicolumn{3}{c|}{CNN~\cite{zhang2017learning} +aug.} & {\bf 31.66} & {\bf 28.26} & {\bf 25.96} \\ \midrule
     \multicolumn{3}{c|}{Original} & 24.82 & 18.97 & 14.52 \\
     \multicolumn{3}{c|}{Random Agents} & 24.67 & 24.00 & 22.58 \\ \midrule
     \multicolumn{3}{c|}{Proposed} & & & \\
     +convGRU & +RMC & +aug. & & & \\
      & & & 31.17 & 27.84 & 25.55 \\
      \checkmark& & & 31.28 & 27.94 & 25.64 \\
      \checkmark&\checkmark & & 31.37 & 27.95 & 25.70 \\
      \checkmark&\checkmark &\checkmark & 31.47 & 28.03 & 25.77 \\
 \bottomrule
  \end{tabular} 
}
\label{tbl:comp_poisson}
\end{table}
We show the comparison of Poisson denoising in~\Tref{tbl:comp_poisson}.
Similar to~\cite{luisier2011image}, we simulated the Poisson noise with different peak intensities.
The lower the peak intensity, the higher is the noise generated.
An almost similar tendency to Gaussian denoising was observed.
The proposed method achieved a slightly lower performance, compared with CNN~\cite{zhang2017learning}.
This is because it is easy to regress the true pixel values and one feedforward pass with CNN~\cite{zhang2017learning} is enough in Gaussian and Poisson denoising.
However, the proposed method achieves better performance than CNN~\cite{zhang2017learning} in more difficult cases such as salt and pepper denoising and image restoration, which are shown later.

\begin{table}[t]
\caption{PSNR [dB] on BSD68 test set with salt and pepper noise.}
\centering
{
{
  \begin{tabular}{ccc|ccc} \toprule
     \multicolumn{3}{c|}{\multirow{2}{*}{Method}} & \multicolumn{3}{c}{Noise density}\\
      & & & 0.1 & 0.5 & 0.9 \\ \toprule
     \multicolumn{3}{c|}{CNN~\cite{zhang2017learning}} & 40.16 & 29.19 & 23.58 \\
     \multicolumn{3}{c|}{CNN~\cite{zhang2017learning} +aug.} & {\bf 40.40} & 29.40 & 23.76 \\ \midrule
     \multicolumn{3}{c|}{Original} & 15.08 & 8.10 & 5.55 \\
     \multicolumn{3}{c|}{Random Agents} & 22.70 & 17.02 & 12.24 \\ \midrule
     \multicolumn{3}{c|}{Proposed} & & & \\
     +convGRU & +RMC & +aug. & & & \\
      & & & 36.51 & 27.91 & 22.73 \\
      \checkmark& & & 37.86 & 29.26 & 23.54 \\
      \checkmark&\checkmark & & 38.46 & 29.78 & 23.78 \\
      \checkmark&\checkmark &\checkmark & 38.82 & {\bf 29.92} & {\bf 23.81} \\
 \bottomrule
  \end{tabular} 
}
}
\label{tbl:comp_sp}
\end{table}
We show the comparison of salt and pepper denoising in~\Tref{tbl:comp_sp}.
We can see that this task is more difficult than Gaussian and Poisson denoising by comparing the PSNR of ``original'' and ``random agents'' in~\Tref{tbl:comp_sp} with those in~\Tref{tbl:comp_gaussian} and~\Tref{tbl:comp_poisson}.
We observed that the RMC significantly improved the performance.
In addition, the proposed method outperformed the CNN~\cite{zhang2017learning} when the noise density is 0.5 and 0.9.
Unlike Gaussian and Poisson noises, it is difficult to regress the noise with one feedforward pass of CNN when the noise density is high because the information of the original pixel value is lost (i.e., the pixel value was changed to 0 or 255 by the noise).
In contrast, the proposed method can predict the true pixel values from the neighbor pixels with the iterative filtering actions.

\begin{figure*}[t]
    \begin{center}
        \subfloat[Gaussian denoising ($\sigma=50$).]{
            \includegraphics[width=0.8\linewidth]{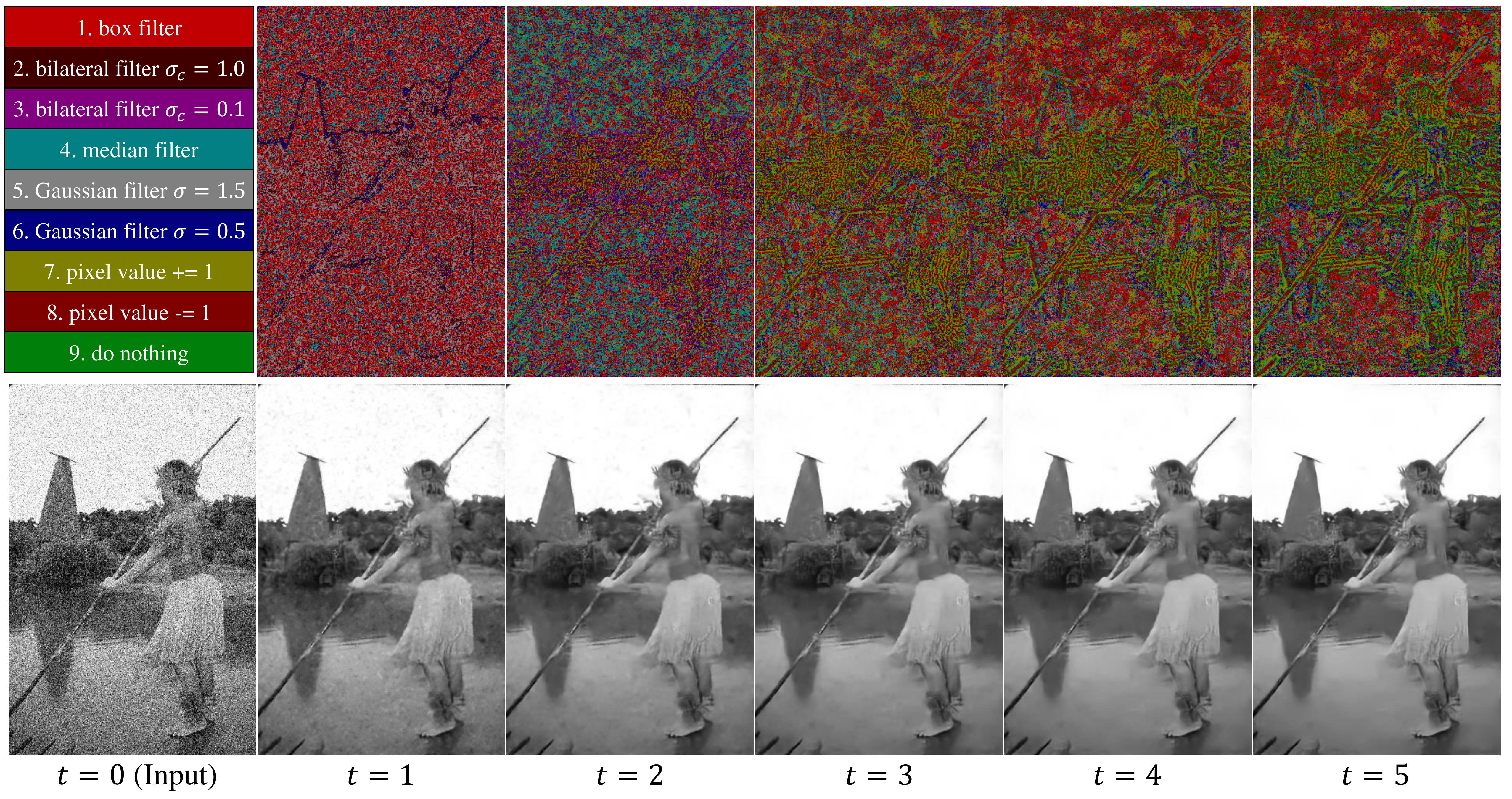}
        }\\
        \subfloat[Salt and pepper denoising (density=0.9).]{
            \includegraphics[width=0.8\linewidth]{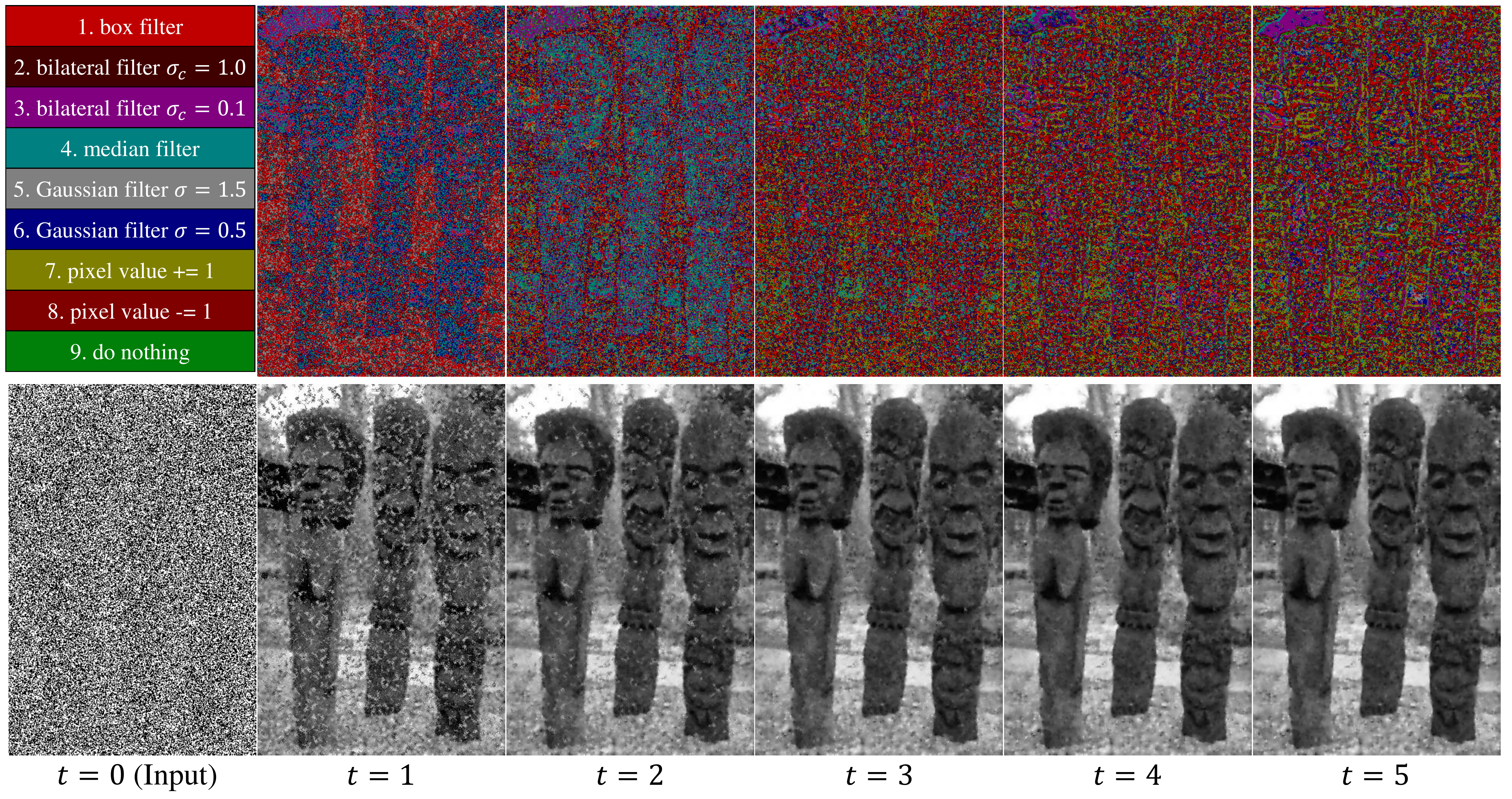}
        }
        \caption{Denoising process of the proposed method and the action map at each time step.}
        \label{fig:vis_denoise}
    \end{center}    
\end{figure*}
We visualize the denoising process of the proposed method, and the action map at each time step in~\Fref{fig:vis_denoise}.
We observed that the noises are iteratively removed by the chosen actions.
In~\Fref{fig:vis_denoise} (a), we can see that the strong filter actions such as box filter are mainly chosen in the flat background regions (sky and water) to remove the noises, and that the foreground object regions (person and bush) are adjusted by the pixel~value~$\pm=$~1 actions.

\begin{figure}[t]
    \begin{center}
        \includegraphics[width=1.0\linewidth]{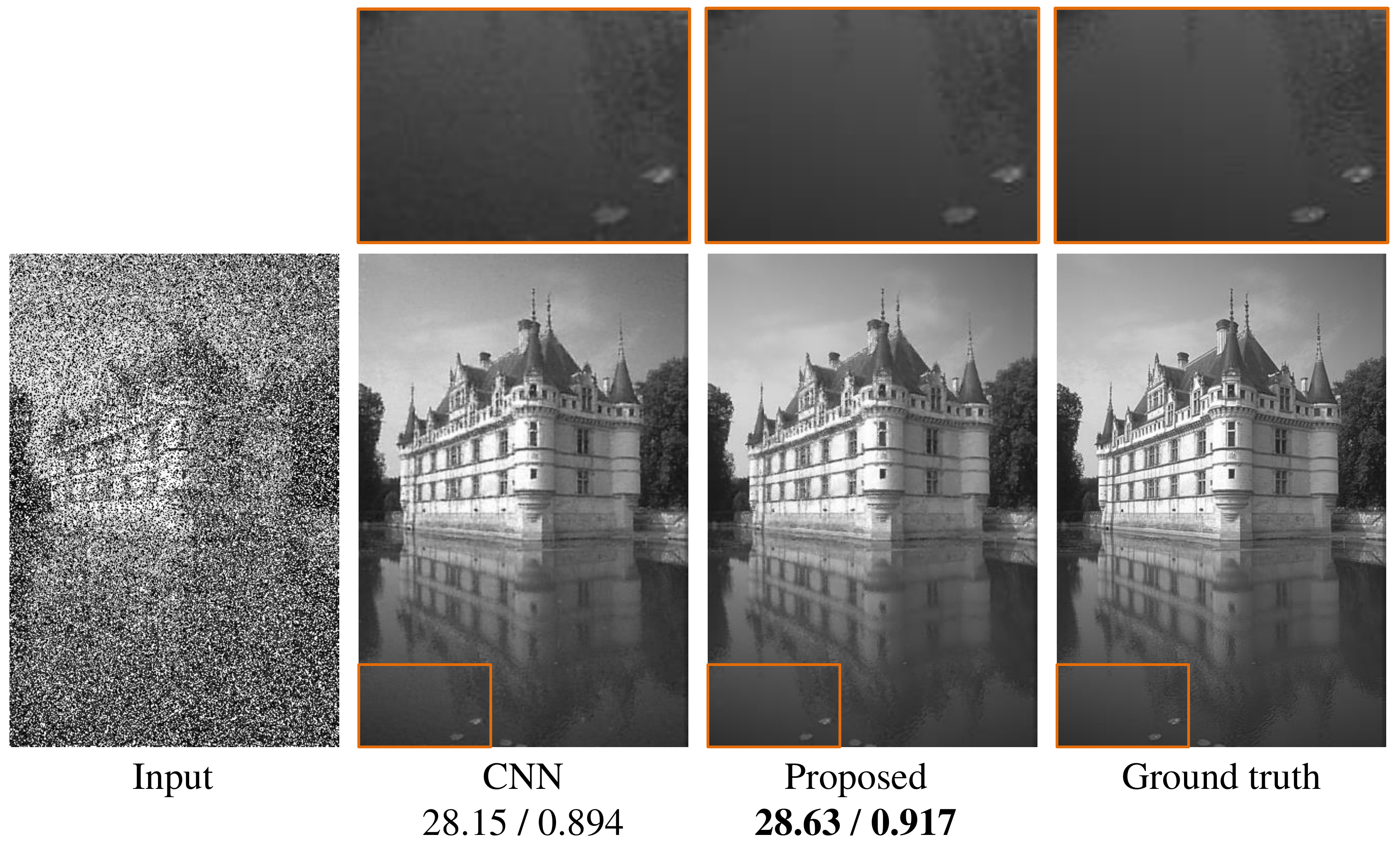}\\
        \includegraphics[width=1.0\linewidth]{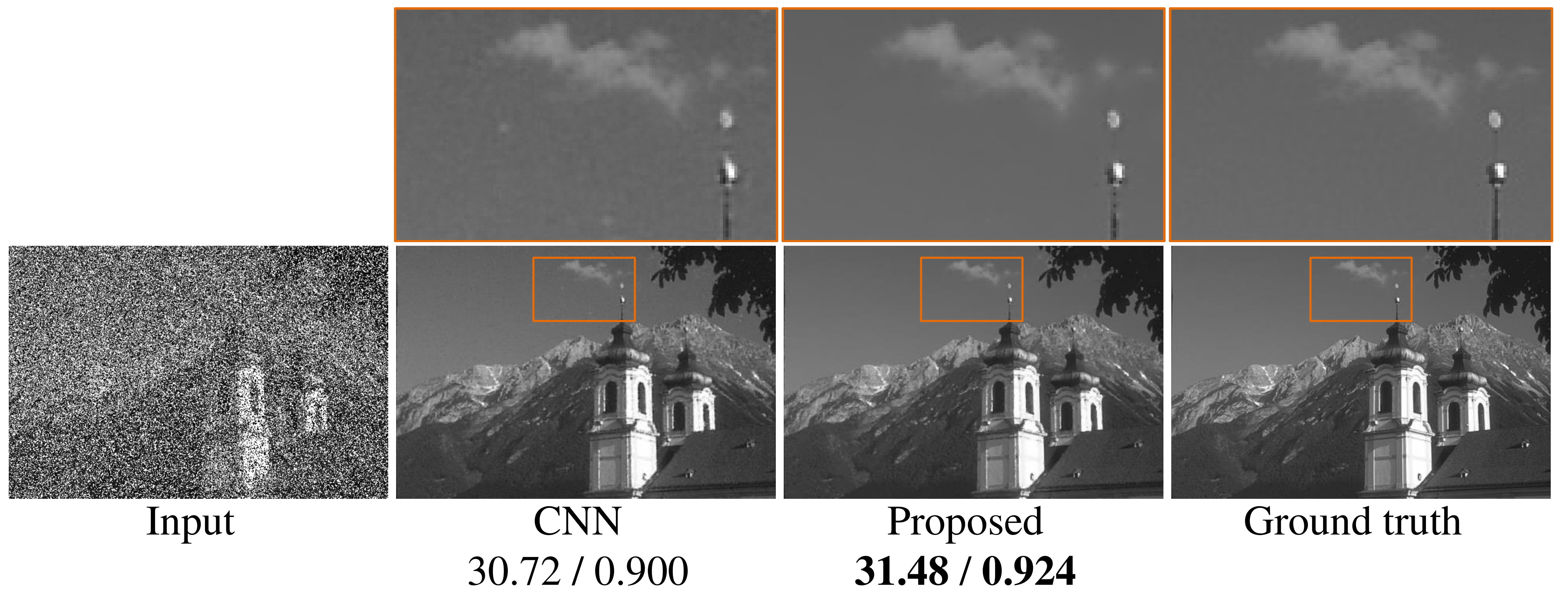}
        \caption{Qualitative comparison of the proposed method and CNN~\cite{zhang2017learning} for salt and pepper noise (density=0.5). PSNR / SSIM are reported.}
        \label{fig:qualitative_denoise}
    \end{center}    
\end{figure}
We show the qualitative comparison with CNN~\cite{zhang2017learning} in~\Fref{fig:qualitative_denoise}.
The proposed method achieved both quantitative and visually better results for salt and pepper denoising.

\subsection{Image Restoration}\label{sec:rest}
\subsubsection{Method}
We applied the proposed method to ``blind'' image restoration, where no mask of blank regions is provided.
The proposed method iteratively inpaints the blank regions by executing actions.
We used the same actions and reward function as those of image denoising, which are shown in~\Tref{tbl:actions_denoise} and~\Eref{eq:reward_denoise}, respectively.

For training, we used 428 training images from the BSD68 train set, 4,774 images from the Waterloo exploration database, and 20,122 images from the ILSVRC2015 val set~\cite{ILSVRC15} (a total of 25,295 images).
We also used 11,343 documents from the Newsgroups 20 train set~\cite{lang1995newsweeder}.
During the training, we created each training image by randomly choosing an image from the 25,295 images and a document from 11,343 documents, and overlaid it on the image.
The font size was randomly decided from the range [10, 30].
The font type was randomly chosen between {\it Arial} and {\it Times New Roman}, where the bold and Italic options were randomly added.
The intensity of the text region was randomly chosen from 0 or 255.
We created the test set that has 68 images by overlaying the randomly chosen 68 documents from the Newsgroup 20 test set on the BSD68 test images.
The settings of the font size and type were the same as those of the training.
The random seed for the test set was fixed between the different methods.
All the hyperparameters were same as those in image denoising, except for the length of the episodes, i.e., $t_{max}=15$.
It required approximately 47 hours for the 30,000 episode training, and 1.5 sec on average for a test image whose size is $481\times 321$.

\subsubsection{Results}\label{sec:restoration_results}
\begin{table}[t]
\caption{Comparison on image restoration.}
\centering
{
{
  \begin{tabular}{cc|cc} \toprule
     \multicolumn{2}{c|}{Method} & PSNR [dB] & SSIM \\ \toprule
     \multicolumn{2}{c|}{Net-D and Net-E~\cite{liu2019deep}} & 29.53 & 0.846 \\
     \multicolumn{2}{c|}{CNN~\cite{zhang2017learning}} & 29.75 & 0.858 \\ \midrule
     \multicolumn{2}{c|}{Original} & 16.61 & 0.656 \\
     \multicolumn{2}{c|}{Random Agents} & 19.58 & 0.476 \\ \midrule
     \multicolumn{2}{c|}{Proposed} & \\
     +convGRU & +RMC & & \\
      \checkmark& & 29.50 & 0.858 \\
      \checkmark&\checkmark & {\bf 29.97} & {\bf 0.868} \\
 \bottomrule
  \end{tabular} 
}
}
\label{tbl:comp_rest}
\end{table}
We show the comparison of the averaged PSNR between the output and ground-truth images in~\Tref{tbl:comp_rest}.
We saved the models of the compared methods at every one epoch, and reported the best results.
For the proposed method, we saved the model at every 300 episodes ($\simeq$ 0.76 epoch) and reported the best results.
The random agents slightly improved PSNR from the original images because the pixel values in the target regions were changed from 0 or 255 by random actions.
However, SSIM of the random agents was significantly worse than the original images because the local patterns in the other regions were broken.
In contrast, the proposed method successfully improved both PSNR and SSIM by learning a strategy for the image restoration.
Here, we compared the proposed method with the two methods (Net-E and Net-D~\cite{liu2019deep} and CNN~\cite{zhang2017learning}) because the Net-E and Net-D achieved much better results than the other restoration methods in the original paper.
We found that the RMC significantly improved the performance, and the proposed method obtained the best result.
This is the similar reason to the case of the salt and pepper noise.
Because the information of the original pixel value is lost by the overlaid texts, its regression is difficult.
In contrast, the proposed method predicts the true pixel value by iteratively propagating the neighbor pixel values with the filtering actions.

\begin{figure*}[t]
    \begin{center}
        \includegraphics[width=0.85\linewidth]{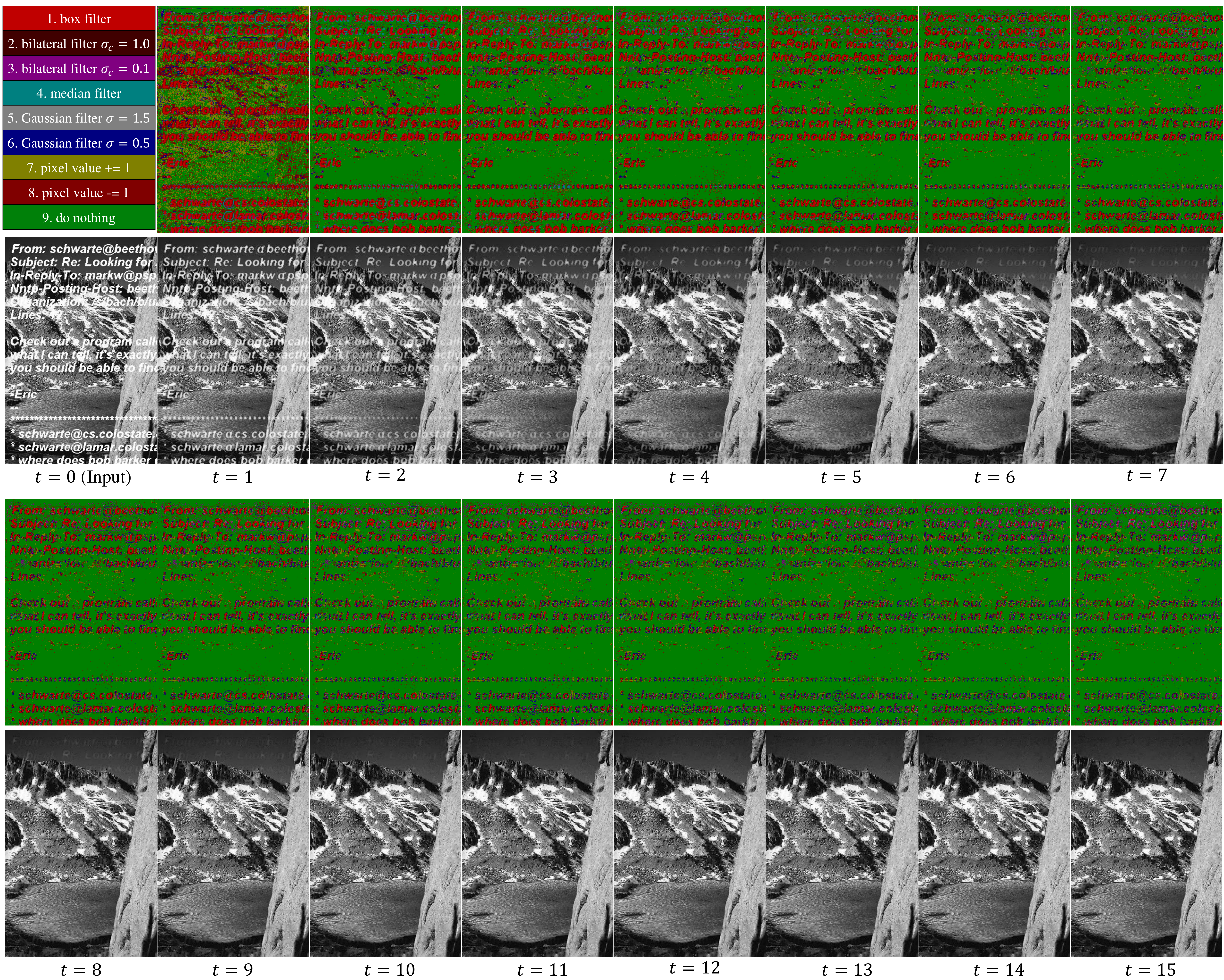}
        \caption{Restoration process of the proposed method and the action map at each time step.}
        \label{fig:vis_rest}
    \end{center}    
\end{figure*}
We visualize the restoration process of the proposed method, and the action map at each time step in~\Fref{fig:vis_rest}.
In the blank regions, the neighbor pixel values are propagated by the strong filter (box filter at $t=1,\cdots,10$) and subsequently by the weak filter (Gaussian filter $\sigma=0.5$ at $t=11,\cdots,15$).
The ``do nothing'' action is chosen in the almost all of the other regions.
\begin{figure*}[t]
    \begin{center}
        \includegraphics[width=0.85\linewidth]{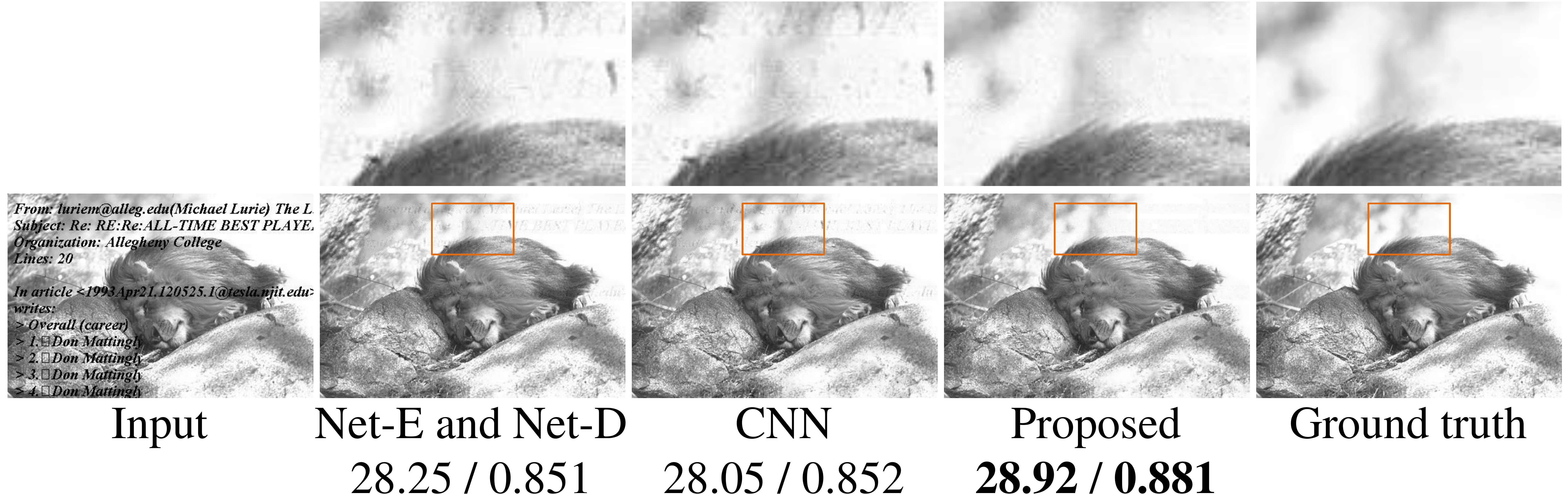}\\
        \includegraphics[width=0.85\linewidth]{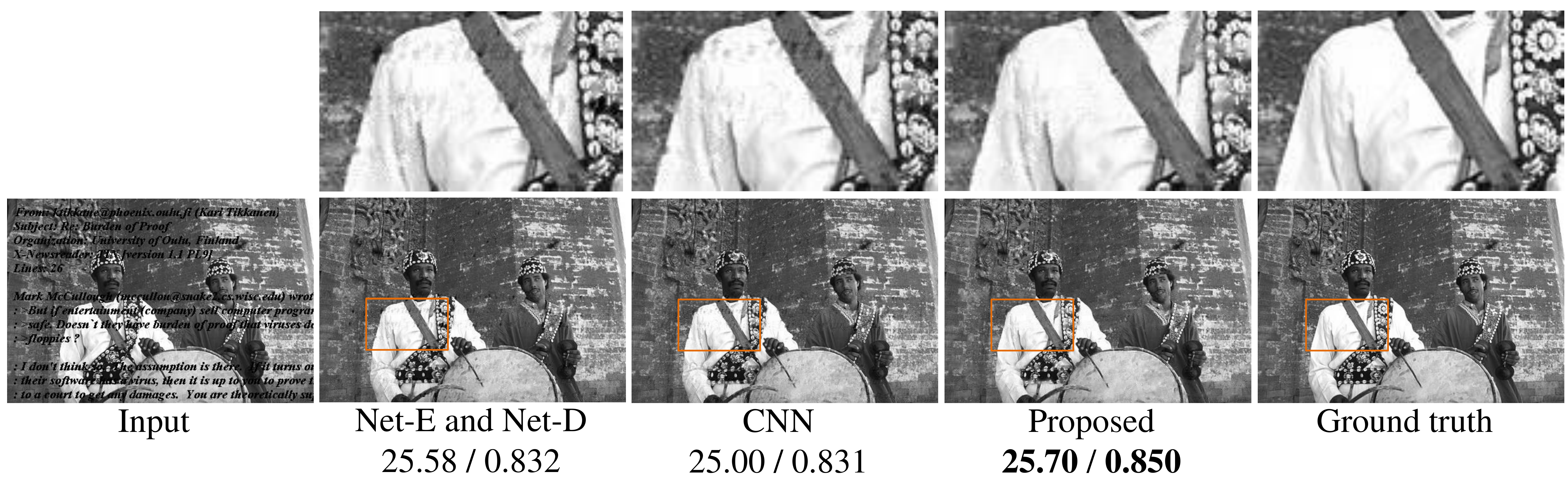}
        \caption{Qualitative comparison of the proposed method with Net-E and Net-D~\cite{liu2019deep} and CNN~\cite{zhang2017learning} on image restoration. PSNR / SSIM are reported.}
        \label{fig:qualitative_rest}
    \end{center}    
\end{figure*}
We show the qualitative comparison with Net-E and Net-D~\cite{liu2019deep} and CNN~\cite{zhang2017learning} in~\Fref{fig:qualitative_rest}.
We observed that there are visually large differences between the results from the proposed method and those from the compared methods.

\begin{figure}[t]
    \begin{center}
        \includegraphics[width=0.9\linewidth]{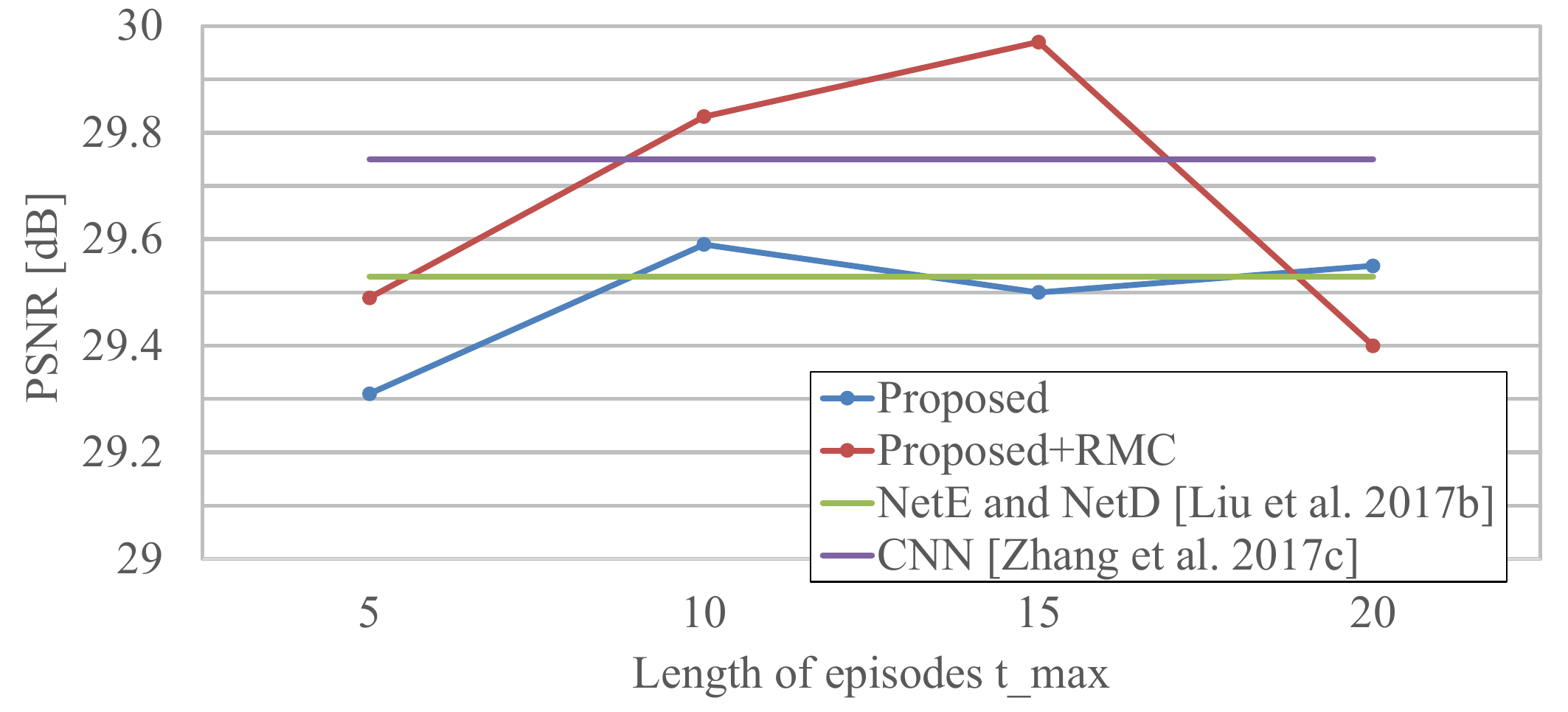}
        \caption{PSNR vs length of episodes ($t_{max}$) on image restoration.}
        \label{fig:length}
    \end{center}    
\end{figure}
We plotted the PSNR vs the length of episodes $t_{max}$ on image restoration in~\Fref{fig:length}.
The performance of the proposed method without the RMC increases when $t_{max}$ gets 5 to 10.
When $t_{max}$ is higher than 10, its performance is almost constant because the agents choose ``pixel value $\pm$ 1'' or ``do nothing'' actions after the missing regions are roughly restored.
We observed that the performance of the proposed method with the RMC increases with the length $t_{max}$ and the peak occurred at $t_{max}=15$.
However, when the length is too long, i.e., $t_{max}=20$, it significantly drops.
This is because it is difficult to consider the far future of neighbor states.
Practically, $t_{max}=10$ is enough because the performance saturates at around 10 iterations.

\subsection{Local Color Enhancement}\label{sec:color}
\subsubsection{Method}
\begin{table}[t]
\caption{Thirteen actions for local color enhancement.}
\centering
{
{
  \begin{tabular}{cccc} \toprule
      \multicolumn{3}{c}{Action} \\ \toprule
     1 / 2 & contrast & $\times0.95$ / $\times1.05$ \\
     3 / 4 & color saturation & $\times0.95$ / $\times1.05$ \\
     5 / 6 & brightness & $\times0.95$ / $\times1.05$ \\
     7 / 8 & red and green & $\times0.95$ / $\times1.05$ \\
     9 / 10 & green and blue & $\times0.95$ / $\times1.05$ \\
     11 / 12 & red and blue & $\times0.95$ / $\times1.05$ \\
     13 & do nothing & \\
 \bottomrule
  \end{tabular} 
}
}
\label{tbl:act_color}
\end{table}
We also applied the proposed method to the local color enhancement.
We used the dataset created by~\cite{yan2016automatic}, which has 70 train images and 45 test images downloaded from Flicker.
Using Photoshop, all the images were enhanced by a professional photographer for three different stylistic local effects: Foreground Pop-Out, Local Xpro, and Watercolor.
Inspired by the prior work for global color enhancement~\cite{jongchan2018distort}, we decided the action set as shown in~\Tref{tbl:act_color}.
We simply chose the same action set as~\cite{jongchan2018distort} except for the ``do nothing'' action. The motivation for the choice has been discussed in the section 4.1 in their paper.
Given an input image $\bm{I}$, the proposed method changes the three channel pixel value at each pixel by executing an action.
When inputting $\bm{I}$ to the network, the RGB color values were converted to CIELab color values.
We defined the reward function as the decrease of L2 distance in the CIELab color space as follows:
\begin{equation}
r_i^{(t)}=||I_i^{target}-s_i^{(t)}||_2-||I_i^{target}-s_i^{(t+1)}||_2.\label{eq:reward_color}
\end{equation}
All the hyperparameters and settings were same as those in image restoration, except for the length of episodes, i.e., $t_{max}=10$.
It required approximately 41 hours for the 30,000 episode training, and 1.6 sec on average for a test image whose size is $384\times 512$.

\subsubsection{Results}
\begin{table}[t]
\caption{Comparison of mean L2 testing errors on local color enhancement. The errors of Lasso, Random Forest, DNN, and Original are from~\cite{yan2016automatic}.}
\centering
{
{
  \begin{tabular}{cc|ccc} \toprule
     \multicolumn{2}{c|}{\multirow{2}{*}{Method}} & Foreground & Local & \multirow{2}{*}{Watercolor} \\
      & & Pop-Out & Xpro &  \\ \toprule
     \multicolumn{2}{c|}{Lasso} & 11.44 & 12.01 & 9.34 \\
     \multicolumn{2}{c|}{Random Forest} & 9.05 & 7.51 & 11.41 \\
     \multicolumn{2}{c|}{DNN~\cite{yan2016automatic}} & 7.08 & 7.43 & 7.20 \\
     \multicolumn{2}{c|}{Pix2pix~\cite{isola2017image}} & {\bf 5.85} & 6.56 & 8.84 \\
     \midrule
     \multicolumn{2}{c|}{Original} & 13.86 & 19.71 & 15.30 \\
     \multicolumn{2}{c|}{Random Agents} & 13.65 & 16.11 & 14.07 \\
     \midrule
     \multicolumn{2}{c|}{Proposed} & & & \\
     +convGRU & +RMC & & & \\
      \checkmark& & 6.75 & 6.17 & 6.44 \\
      \checkmark&\checkmark & 6.69 & {\bf 5.67} & {\bf 6.41} \\
 \bottomrule
  \end{tabular} 
}
}
\label{tbl:comp_color}
\end{table}
We show the comparison of mean L2 errors on 45 test images in~\Tref{tbl:comp_color}.
The proposed method achieved better results than DNN~\cite{yan2016automatic} on all three enhancement styles, and comparable or slightly better results than pix2pix.
We observed that the RMC improved the performance although their degrees of improvement depended on the styles.
It is noteworthy that the existing color enhancement method using deep RL~\cite{jongchan2018distort,hu2017exposure} cannot be applied to this local enhancement application because they can execute only global actions.

\begin{figure*}[t]
    \begin{center}
        \includegraphics[width=0.85\linewidth]{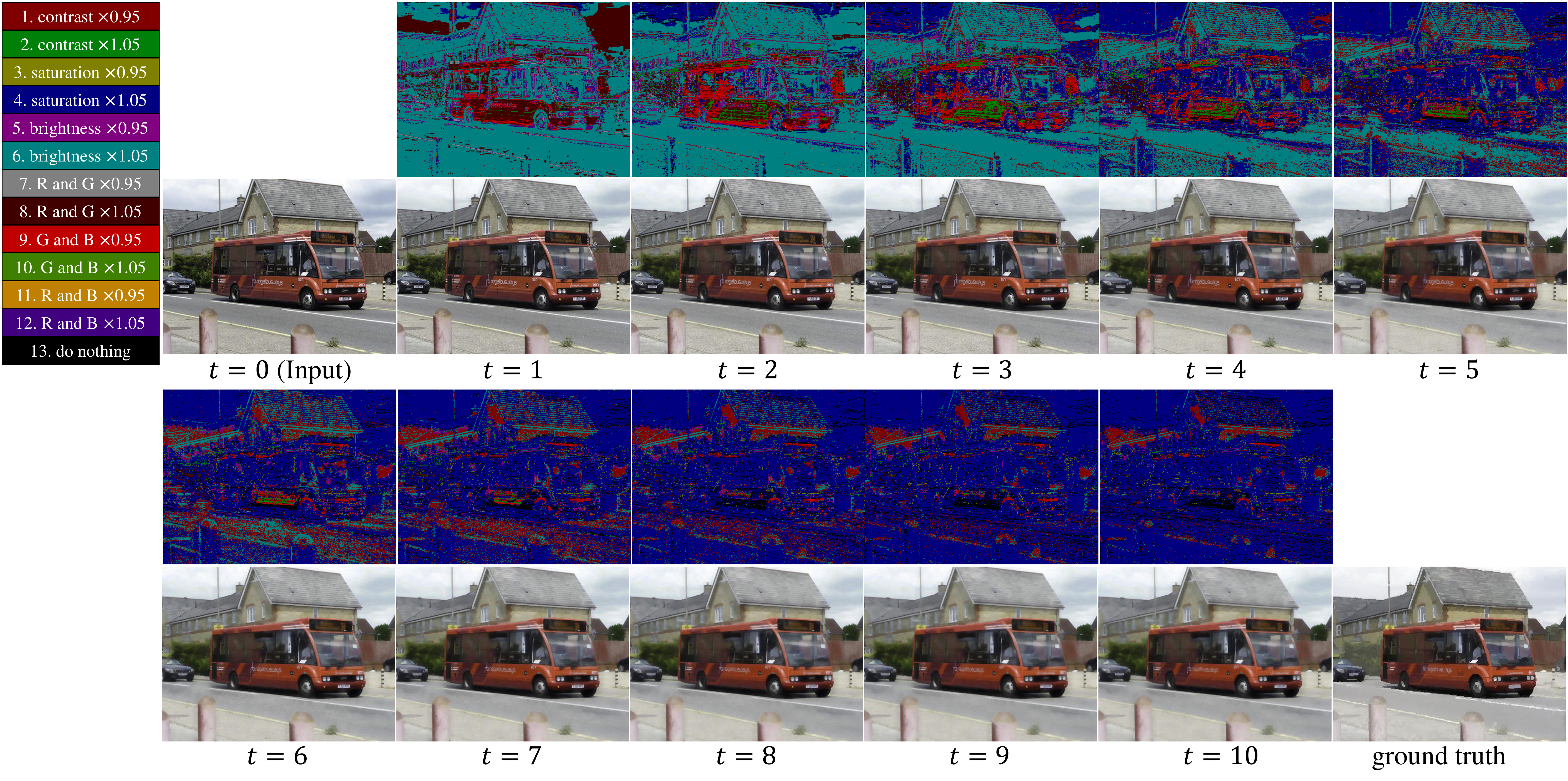}
        \caption{Color enhancement process of the proposed method for watercolor, and the action map at each time step.}
        \label{fig:vis_color}
    \end{center}    
\end{figure*}
We visualize the color enhancement process of the proposed method, and the action map at each time step in~\Fref{fig:vis_color}.
Similar to~\cite{jongchan2018distort}, the proposed method is interpretable while the DNN-based color mapping method~\cite{yan2016automatic} is not.
We can see that the brightness and saturation were mainly increased to convert the input image to watercolor style.

\begin{figure}[t]
    \begin{center}
        \includegraphics[width=1.0\linewidth]{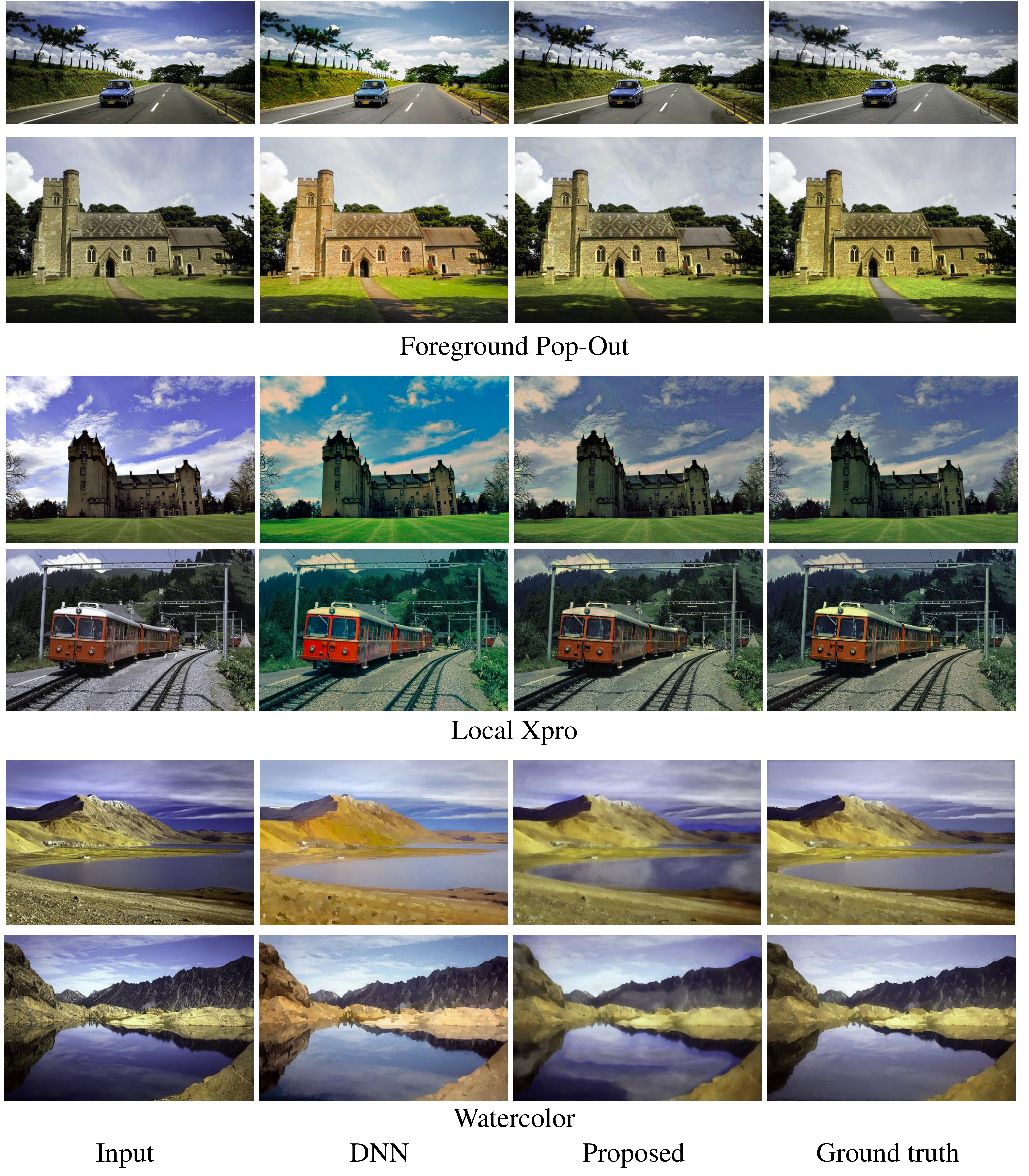}
        \caption{Qualitative comparison of the proposed method and DNN~\cite{yan2016automatic}. The saturation of the images from DNN appear higher owning to the color correction for the sRGB space (for details, see~\protect\url{https://github.com/stephenyan1231/dl-image-enhance}).}
        \label{fig:qualitative_color}
    \end{center}    
\end{figure}
We show the qualitative comparison between the proposed method and DNN~\cite{yan2016automatic} in~\Fref{fig:qualitative_color}.
The proposed method achieved both quantitatively and qualitatively better results.

\subsection{Saliency-Driven Image Editing}
\subsubsection{Method}
We applied the proposed method to saliency-driven object enhancement, which is one of the applications of saliency-driven image editing proposed in~\cite{mechrez2018saliency}.
Input is a set of a color image $\bm{I}$ and a mask image $\bm{M}$ that shows target regions.
The objective of this task is to increase the saliency values in the target regions and decrease the saliency values in the other regions while keeping the naturalness of the image.
We used a dataset created by~\cite{mechrez2018saliency}, which contains 101 images and corresponding masks.
Our network takes a four channel image (concatenation of the color and mask images) as input, and the agents edit the color image by iteratively executing actions.
We used the same action set as local color enhancement in~\Tref{tbl:act_color}.
We defined the reward function as follows:
\begin{equation}
r_i^{(t)}=
\begin{cases}
\alpha(S_i(\bm{s}^{(t+1)})-S_i(\bm{s}^{(t)})) &\text{if} \ M_i=255\\
-\beta(S_i(\bm{s}^{(t+1)})-S_i(\bm{s}^{(t)})) &\text{if} \ M_i=0,
\end{cases}\label{eq:reward_saliency}
\end{equation}
where $S_i(\bm{s}^{(t)})$ is the saliency value of the current image $\bm{s}^{(t)}$ at $i$-th pixel, and $M_i$ is the pixel value at the $i$-th pixel in the mask image.
We set $\alpha=1.0$ and $\beta=0.5$.
\Eref{eq:reward_saliency} means that the agents in/outside the target regions can get positive rewards when the saliency values increase/decrease.
It is noteworthy that this application does not need ground truth images (i.e., unsupervised learning) in contrast to image denoising. image restoration, and local color enhancement.
We used the saliency map estimation method in~\cite{montabone2010human} to compute $S_i(\bm{s}^{(t)})$, which is implemented in OpenCV library.
We initialized the policy and value networks with the weights trained on local color enhancement except for the last layers and convGRU.
Given an input (a color image and a mask), we train the networks to maximize the reward in~\Eref{eq:reward_saliency} using the input (i.e., minibatch size is 1).
We test the networks after 200 episode training for each input.
We did not perform random cropping and rotation during the training.
When testing, we used the guided filter~\cite{he2010guided} to smooth the edited color image at each time step.
It required 364 sec for the 200 episode training for an image whose size is $512\times 384$, and 1.5 sec for testing on a single Tesla V100 GPU when $t_{max}=10$.

\subsubsection{Results}
\begin{figure*}[t]
    \begin{center}
        \includegraphics[width=0.85\linewidth]{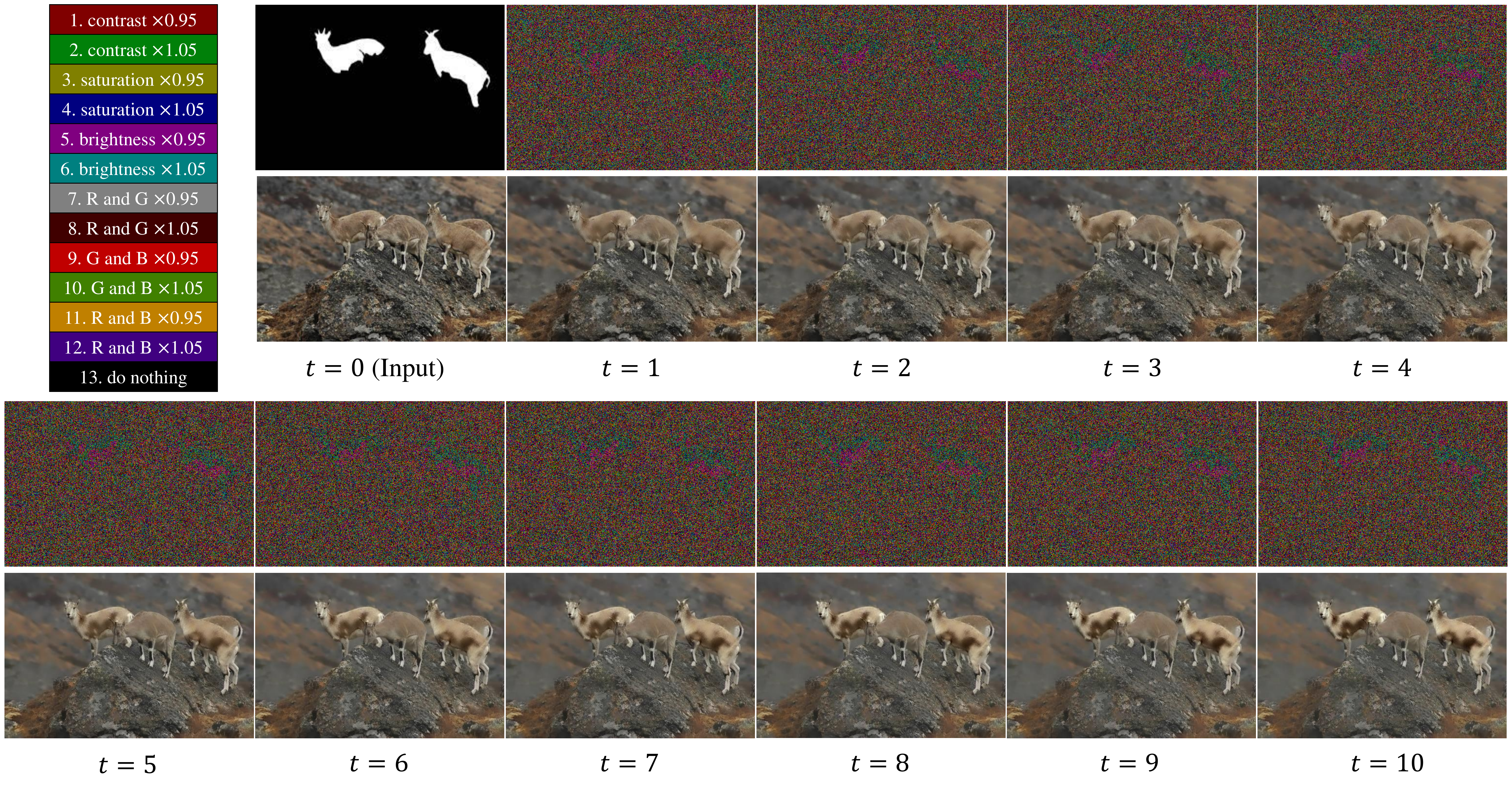}
        \caption{Saliency-driven image editing process of the proposed method and the action map at each time step.}
        \label{fig:vis_saliency}
    \end{center}    
\end{figure*}
We visualize the saliency-driven image editing process of the proposed method and the action map at each time step in~\Fref{fig:vis_saliency}.
We observed that the appearances of the target deer are gradually emphasized by ``brightness$\times$1.05'' and ``brightness$\times$0.95'' actions.

\begin{figure*}[t]
    \begin{center}
        \includegraphics[width=1.0\linewidth]{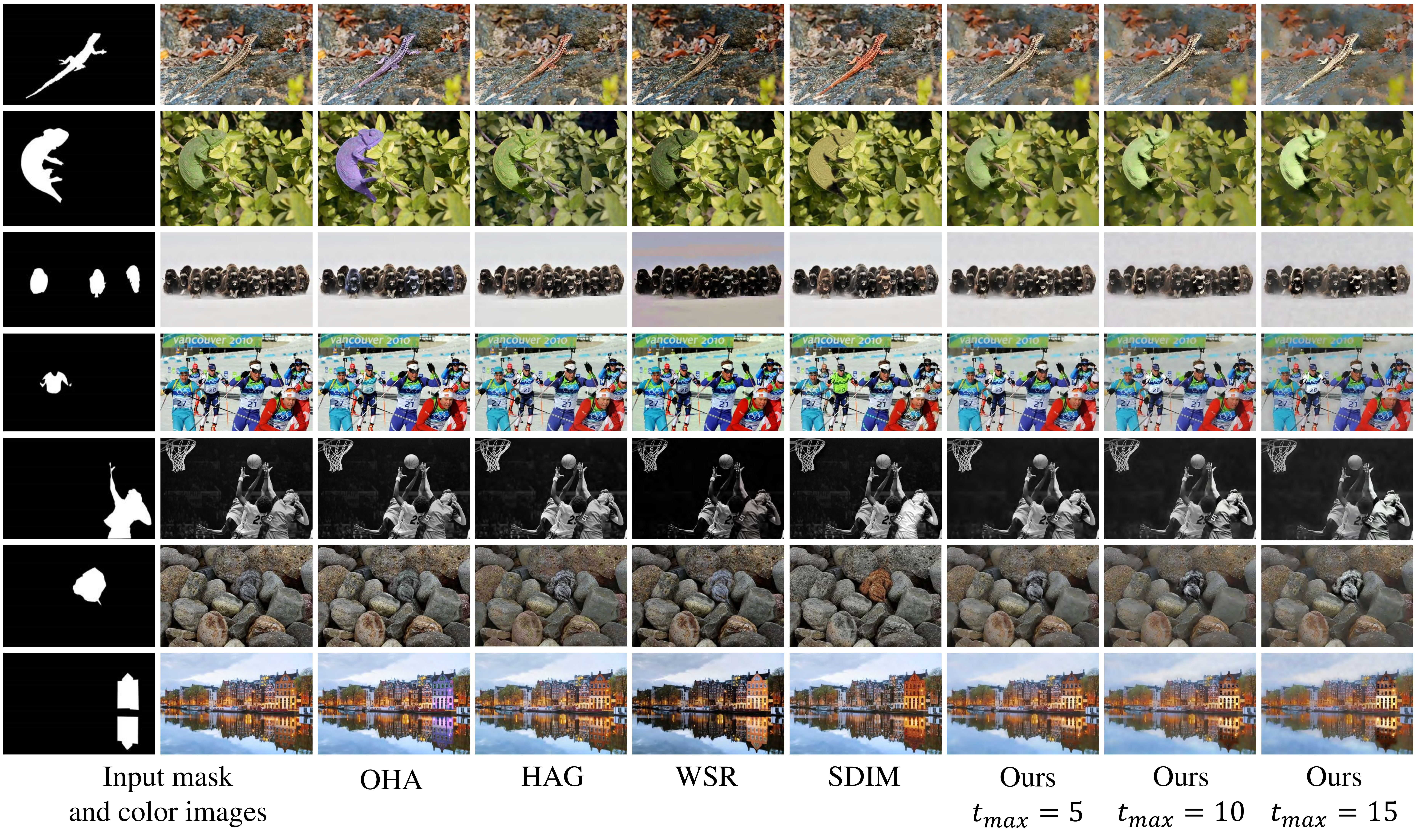}
        \caption{Qualitative comparison on saliency-driven object enhancement.}
        \label{fig:comp_saliency}
    \end{center}    
\end{figure*}
We show the qualitative comparisons with other saliency-driven image editing methods (OHA~\cite{mateescu2014attention}, HAG~\cite{hagiwara2011saliency}, WSR~\cite{wong2011saliency}, and SDIM~\cite{mechrez2018saliency}) in~\Fref{fig:comp_saliency}.
We observed that our method can successfully enhance the target regions and diminish the other regions while keeping the naturalness of the image.
In our method, we can control how much the input color image is edited by changing the length of each episode ($t_{max}$).

\begin{figure*}[t]
    \begin{center}
        \includegraphics[width=1.0\linewidth]{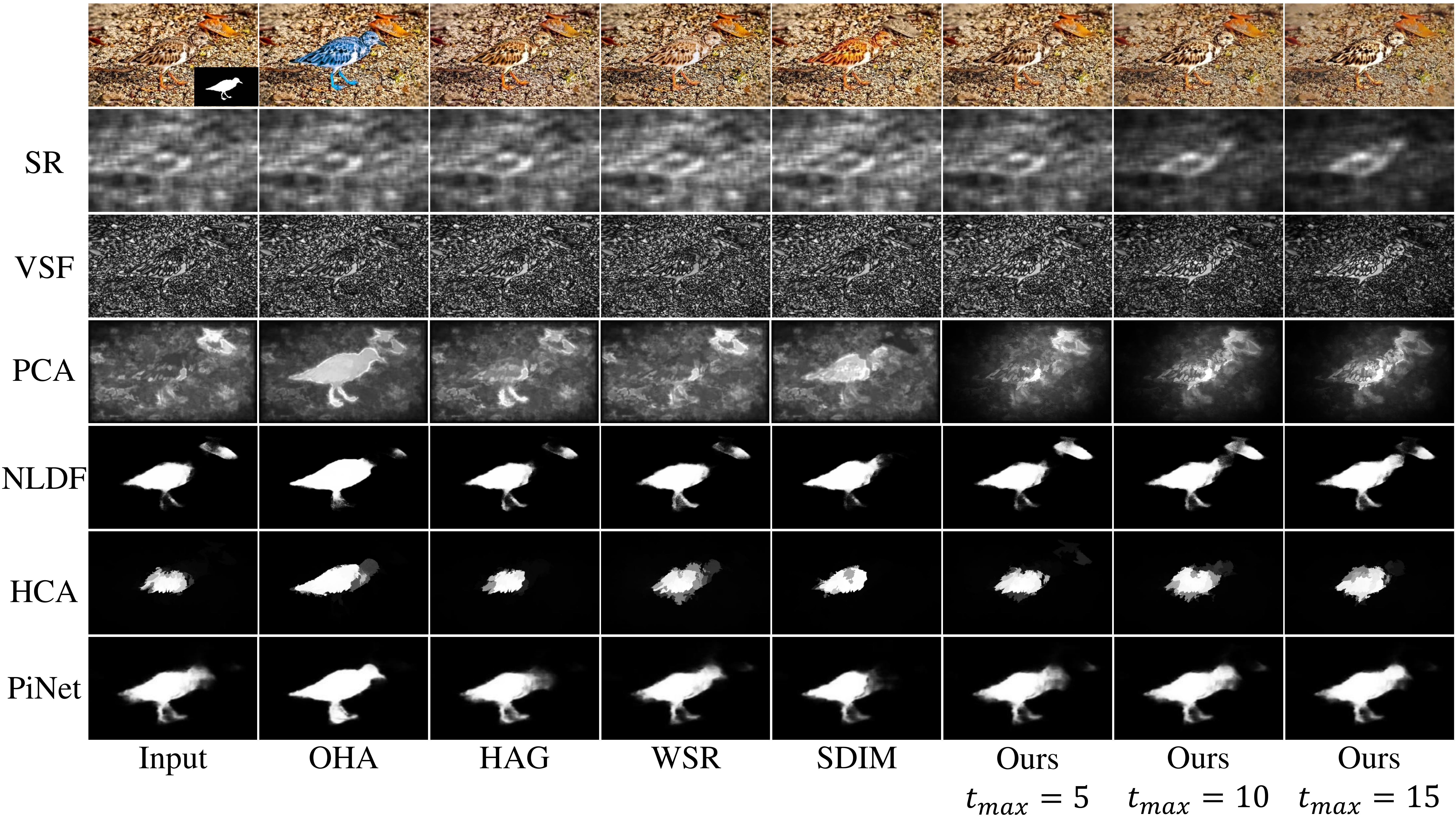}
        \caption{Visual comparison on different saliency map estimation methods.}
        \label{fig:comp_saliencymaps}
    \end{center}    
\end{figure*}
We show the saliency maps of the outputs from each saliency-driven image editing method in~\Fref{fig:comp_saliencymaps}.
We used six saliency map estimation methods including both hand-crafted ones (SR~\cite{hou2007saliency}, VSF~\cite{montabone2010human}, and PCA~\cite{margolin2013makes}) and deep learning-based ones (NLDF~\cite{luo2017non}), HCA~\cite{qin2018hierarchical}, and PiCANet~\cite{liu2018picanet}).
The proposed method successfully learns how to enhance/diminish the target/non-target regions.
That was confirmed by a wide variety of saliency map estimation methods.

\begin{table*}[t]
\caption{Comparison of Pearson correlation coefficient (CC) and structural similarity (SSIM) on saliency-driven object enhancement.}
\centering
{
{
  \begin{tabular}{cc|cccccc|c} \toprule
     \multicolumn{2}{c|}{\multirow{2}{*}{Method}} & \multicolumn{6}{|c|}{CC} & \multirow{2}{*}{SSIM} \\
	      & & SR~\cite{hou2007saliency} & VSF~\cite{montabone2010human} & PCA~\cite{margolin2013makes}& NLDF~\cite{luo2017non} & HCA~\cite{qin2018hierarchical} & PiCANet~\cite{liu2018picanet} &  \\ \toprule
     \multicolumn{2}{c|}{OHA~\cite{mateescu2014attention}} & 0.23 & 0.16 & 0.48 & 0.63 & 0.59 & 0.61 & 0.98 \\
     \multicolumn{2}{c|}{HAG~\cite{hagiwara2011saliency}} & 0.29 & 0.20 & 0.50 & 0.64 & 0.57 & 0.62 & 0.95 \\
     \multicolumn{2}{c|}{WSR~\cite{wong2011saliency}} & 0.25 & 0.17 & 0.42 & 0.62 & 0.56 & 0.61 & 0.86 \\
     \multicolumn{2}{c|}{SDIM~\cite{mechrez2018saliency}} & 0.23 & 0.15 & {\bf 0.62} & 0.65 & 0.58 & {\bf 0.64} & 0.96 \\
     \midrule
     \multicolumn{2}{c|}{Original} & 0.24 & 0.16 & 0.37 & 0.61 & 0.53 & 0.60 & 1.0 \\
     \multicolumn{2}{c|}{Random Agents ($t_{max}=15$)} & 0.23 & 0.16 & 0.43 & 0.61 & 0.52 & 0.58 & 0.90 \\
     \midrule
     \multicolumn{2}{c|}{Proposed ($t_{max}=5$)} & & & & & & & \\
     +convGRU & +RMC & & & & & & \\
      \checkmark& & 0.37 & 0.30 & 0.48 & 0.63 & 0.54 & 0.60 & 0.89 \\
      \checkmark&\checkmark & 0.39 & 0.32 & 0.49 & 0.63 & 0.55 & 0.60 & 0.88 \\
     \midrule
     \multicolumn{2}{c|}{Proposed ($t_{max}=10$)} & & & & & & & \\
     +convGRU & +RMC & & & & & & \\
      \checkmark& & 0.49 & 0.45 & 0.54 & 0.65 & 0.58 & 0.62 & 0.82 \\
      \checkmark&\checkmark & 0.48 & 0.42 & 0.52 & 0.65 & 0.57 & 0.62 & 0.77 \\
     \midrule
     \multicolumn{2}{c|}{Proposed ($t_{max}=15$)} & & & & & & & \\
     +convGRU & +RMC & & & & & & \\
      \checkmark& & {\bf 0.54} & {\bf 0.50} & 0.57 & {\bf 0.67} & 0.58 & {\bf 0.64} & 0.77 \\
      \checkmark&\checkmark & 0.53 & 0.49 & 0.56 & {\bf 0.67} & {\bf 0.59} & 0.62 & 0.73 \\
 \bottomrule
  \end{tabular} 
}
}
\label{tbl:comp_saliency}
\end{table*}
Similar to~\cite{mechrez2018saliency}, we regarded the input mask as a groundtruth saliency map, and computed Pearson correlation coefficient (CC) between the input mask and the saliency map of output color image.
We show the average CC on 101 images of each saliency-driven image editing method in~\Tref{tbl:comp_saliency}.
Although the random agents no longer work on this application, the proposed method successfully learned a strategy for increasing the saliency values in the target regions and decreasing the saliency values in the other regions.
SDIM~\cite{mechrez2018saliency} achieved the best result on PCA~\cite{margolin2013makes} because it used the saliency maps from PCA~\cite{margolin2013makes} in its image editing algorithm.
However, the proposed method obtained the best results on the other saliency map estimation methods.
We observed that the RMC does not always boost the performance.
That may be because this application is unsupervised learning (i.e., no ground truth images) and the reward values are not stable in contrast to the other three applications.

To evaluate the naturalness of the output images, we also report the average structural similarity (SSIM) between the output color images and original color images.
In terms of SSIM, our method is lower than the other methods because our method changes every pixel value although the other methods except for WSR edit only a part of the input image.
However, from~\Fref{fig:comp_saliency}, we can observe that the output images from our method are as natural as those by the other methods.

\section{Conclusions}\label{sec:conclusions}
We proposed a novel pixelRL problem setting and applied it to four different applications: image denoising, image restoration, local color enhancement, and saliency-driven image editing.
We also proposed an effective learning method for the pixelRL problem, which boosts the performance of the pixelRL agents.
Our experimental results demonstrated that the proposed method achieved comparable or better results than state-of-the-art methods on each application.
Different from the existing deep learning-based methods for such applications, the proposed method is interpretable.
The interpretability of deep learning has been attracting much attention~\cite{selvaraju2017grad,zeiler2014visualizing}, and it is especially important for some applications such as medical image processing~\cite{razzak2018deep}.

The proposed method can maximize the pixel-wise reward; in other words, it can minimize the pixel-wise non-differentiable objective function such as hand-crafted saliency values.
That is a completely different point from conventional CNN-based image processing methods.
Therefore, we believe that the proposed method can be potentially used for more image processing applications where supervised learning cannot be applied.

\section*{Acknowledgments}
This work was partially supported by the Grants-in-Aid for Scientific Research (no. 16J07267) from JSPS and JST-CREST (JPMJCR1686).


%

\begin{IEEEbiography}[{\includegraphics[width=1in,height=1.25in,clip,keepaspectratio]{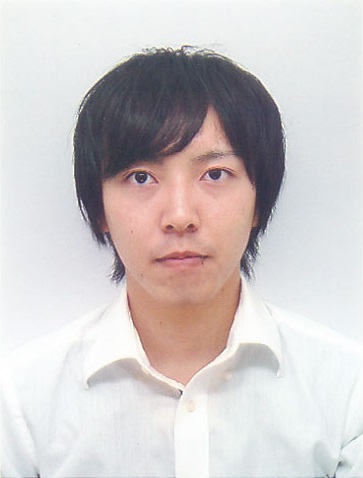}}]{Ryosuke Furuta}
received the B.S., M.S., and Ph.D. degrees in information and communication engineering from The University of Tokyo, Tokyo, Japan, in 2014, 2016 and 2019, respectively.
He is currently an Assistant Professor at the Department of Information and Computer Technology, Tokyo University of Science.
His current research interests include computer vision, machine learning, and image processing, especially MRF optimization. 

Dr. Furuta is a member of IEEE, ACM, AAAI and ITE. 
\end{IEEEbiography}

\begin{IEEEbiography}[{\includegraphics[width=1in,height=1.25in,clip,keepaspectratio]{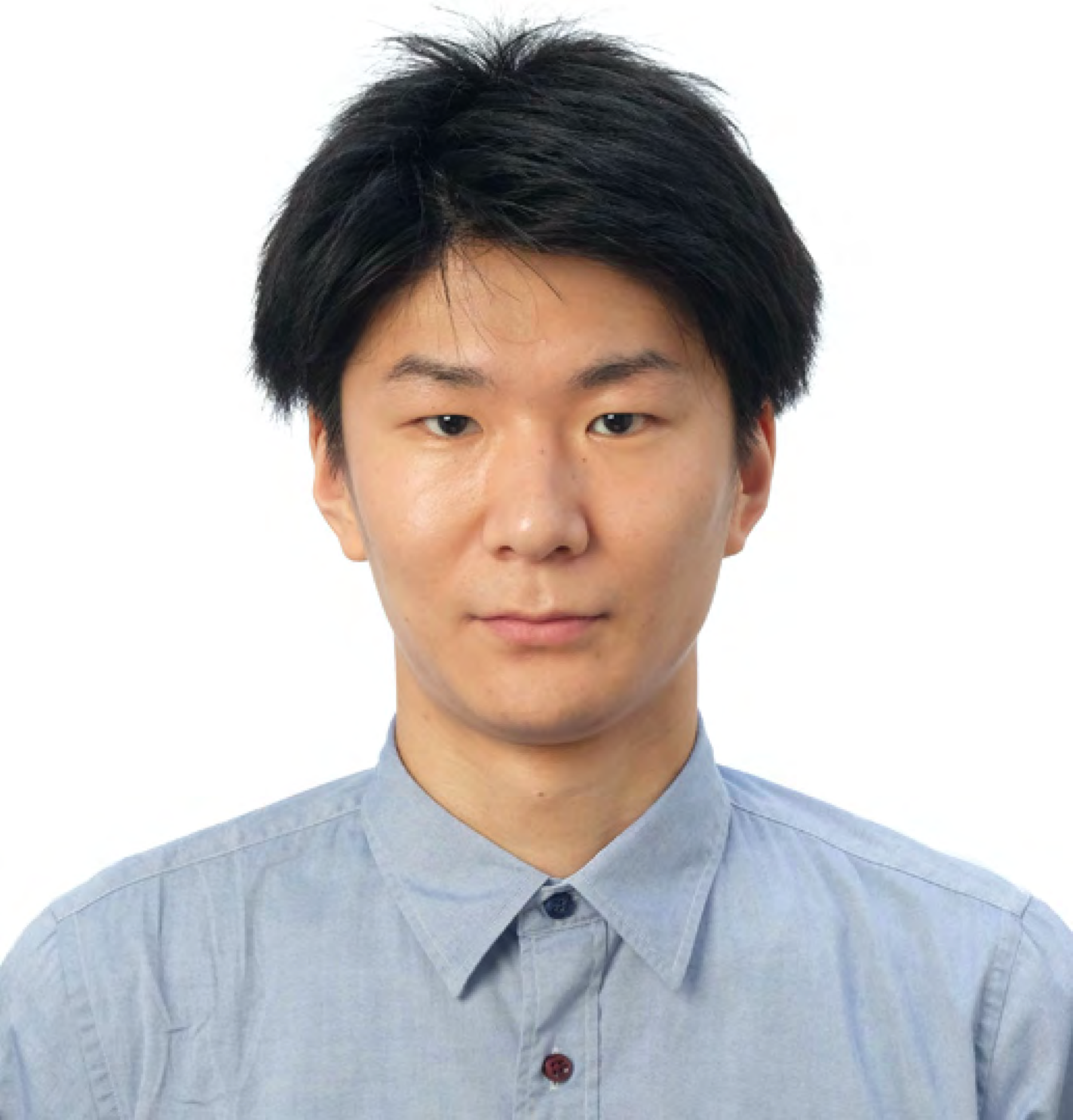}}]{Naoto Inoue}
received the B.E. and M.S. in information and communication engineering from the University of Tokyo in 2016 and 2018, respectively. 
He is currently a Ph.D. student in The University of Tokyo, Japan. 
His research interests lie in computer vision, with particular interest in object detection and domain adaptation. 
He is a member of IEEE.
\end{IEEEbiography}

\begin{IEEEbiography}[{\includegraphics[width=1in,height=1.25in,clip,keepaspectratio]{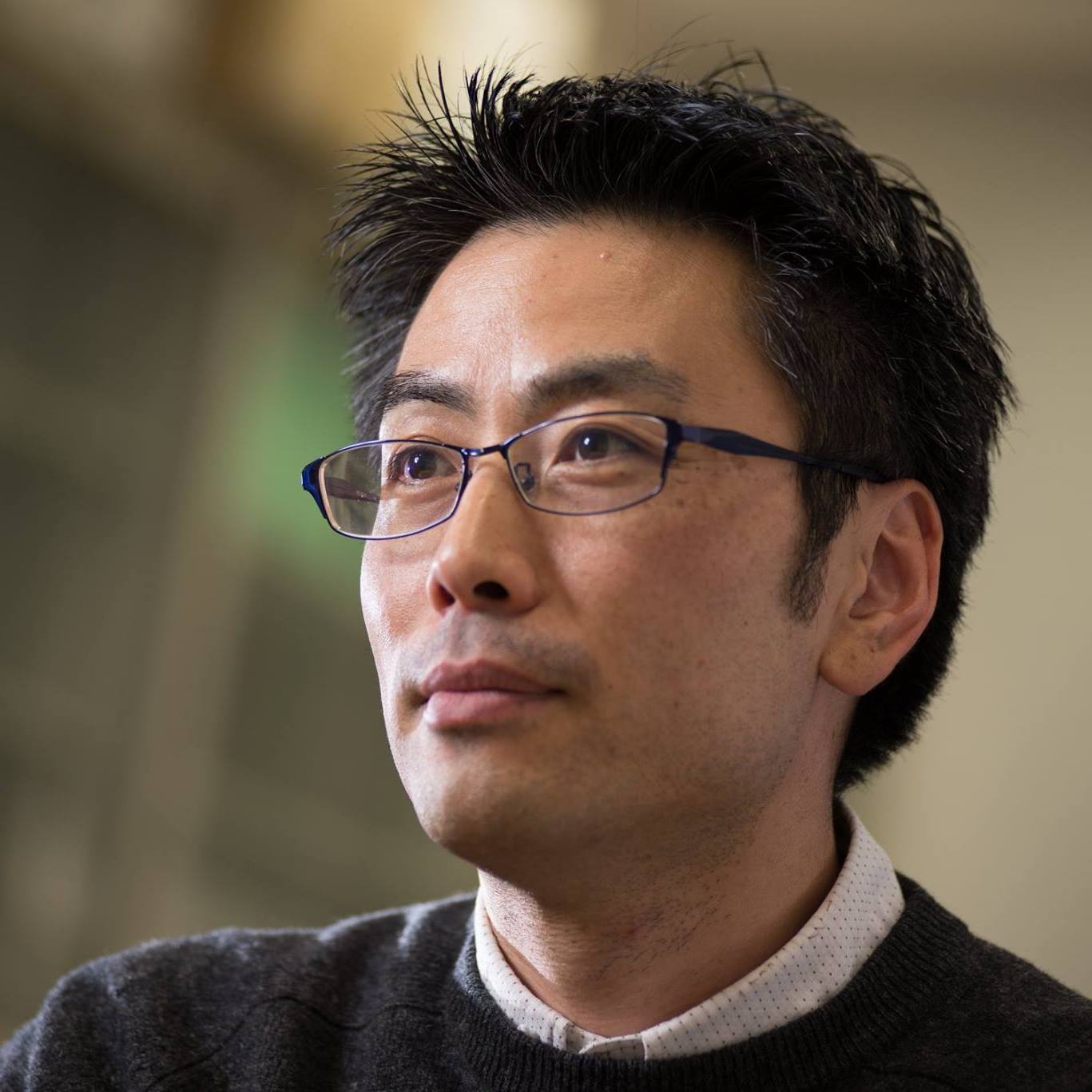}}]{Toshihiko Yamasaki}
received the B.S. degree in electronic engineering, the M.S. degree in information and communication engineering, and the Ph.D. degree from The University of Tokyo in 1999, 2001, and 2004, respectively. 
He is currently an Associate Professor at Department of Information and Communication Engineering, Graduate School of Information Science and Technology, The University of Tokyo. 
He was a JSPS Fellow for Research Abroad and a visiting scientist at Cornell University from Feb. 2011 to Feb. 2013.
His current research interests include attractiveness computing based on multimedia big data analysis, pattern recognition, machine learning, and so on.

Dr. Yamasaki is a member of IEEE, ACM, AAAI, IEICE, ITE, and IPSJ.
\end{IEEEbiography}




\end{document}